\documentclass[sigconf]{acmart}  %

\AtBeginDocument{%
  \providecommand\BibTeX{{%
    \normalfont B\kern-0.5em{\scshape i\kern-0.25em b}\kern-0.8em\TeX}}}

\copyrightyear{2024} 
\acmYear{2024} 
\setcopyright{rightsretained} 
\acmConference[E-Energy '24]{The 15th ACM International Conference on Future and Sustainable Energy Systems}{June 4--7, 2024}{Singapore, Singapore}
\acmBooktitle{The 15th ACM International Conference on Future and Sustainable Energy Systems (E-Energy '24), June 4--7, 2024, Singapore, Singapore}\acmDOI{10.1145/3632775.3639589}
\acmISBN{979-8-4007-0480-2/24/06}

\usepackage{xcolor}
\usepackage{subcaption}

\usepackage{booktabs}
\usepackage{multirow}
\usepackage{colortbl}

\usepackage{algorithm}
\usepackage[noend]{algpseudocode}
\usepackage{hyperref}

\usepackage{todonotes}
\setuptodonotes{inline}

\newcommand{\LineComment}[1]{\State {\small\color{teal} \# #1}}

\newcommand*\circled[1]{\tikz[baseline=(char.base)]{\node[shape=circle,draw,inner sep=1pt] (char) {#1};}}

\newcommand{\parag}[1]{\vspace{2mm plus 1mm minus 1mm}\noindent\textbf{#1.}\hspace{2mm}}

\begin{document}

\title{FedZero: Leveraging Renewable Excess Energy in Federated~Learning}

\author{Philipp Wiesner}
\email{wiesner@tu-berlin.de}
\orcid{0000-0001-5352-7525}
\affiliation{%
  \institution{TU Berlin}
  \country{}
}

\author{Ramin Khalili}
\email{ramin.khalili@huawei.com}
\orcid{0000-0003-2463-7033}
\affiliation{%
  \institution{Huawei}
  \country{}
}

\author{Dennis Grinwald}
\email{dennis.grinwald@tu-berlin.de}

\orcid{0000-0001-9903-2886}
\affiliation{%
  \institution{TU Berlin}
  \country{}
}
\author{Pratik Agrawal}
\email{pratik.agrawal@tu-berlin.de}
\orcid{0009-0000-8294-9823}
\affiliation{%
  \institution{TU Berlin}
  \country{}
}

\author{Lauritz Thamsen}
\email{lauritz.thamsen@glasgow.ac.uk}
\orcid{0000-0003-3755-1503}
\affiliation{%
  \institution{University of Glasgow}
  \country{}
}

\author{Odej Kao}
\email{odej.kao@tu-berlin.de}
\orcid{0000-0001-6454-6799}
\affiliation{%
  \institution{TU Berlin}
  \country{}
}

\renewcommand{\shortauthors}{Wiesner et al.}

\begin{abstract}
Federated Learning (FL) is an emerging machine learning technique that enables distributed model training across data silos or edge devices without data sharing.
Yet, FL inevitably introduces inefficiencies compared to centralized model training, which will further increase the already high energy usage and associated carbon emissions of machine learning in the future.
One idea to reduce FL's carbon footprint is to schedule training jobs based on the availability of renewable excess energy that can occur at certain times and places in the grid.
However, in the presence of such volatile and unreliable resources, existing FL schedulers cannot always ensure fast, efficient, and fair training.

We propose FedZero, an FL system that operates exclusively on renewable excess energy and spare capacity of compute infrastructure to effectively reduce a training's operational carbon emissions to zero.
Using energy and load forecasts, FedZero leverages the spatio-temporal availability of excess resources by selecting clients for fast convergence and fair participation.
Our evaluation, based on real solar and load traces, shows that FedZero converges significantly faster than existing approaches under the mentioned constraints while consuming less energy.
Furthermore, it is robust to forecasting errors and scalable to tens of thousands of clients.
\end{abstract}

\ccsdesc[500]{Social and professional topics~Sustainability}
\ccsdesc[300]{Computing methodologies~Distributed artificial intelligence}

\keywords{sustainable computing, carbon efficiency, electricity curtailment, federated learning, client selection, green AI}

\maketitle

\section{Introduction}
\label{sec:intro}

\begin{figure}[t]
    \centering
    \hspace{-4mm}
    \includegraphics[width=0.85\columnwidth]{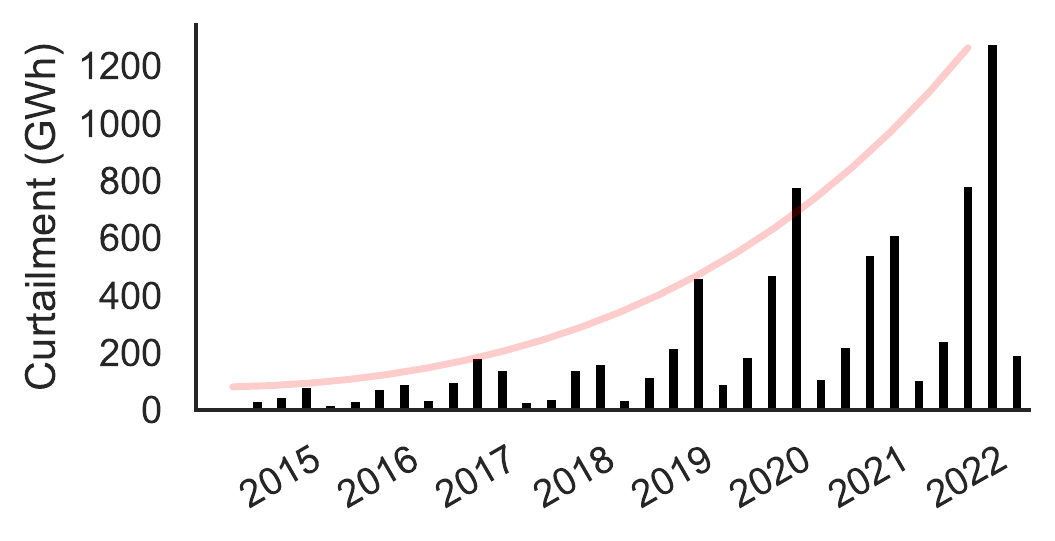}
    \vspace{-4mm}
    \caption{Quarterly wind and solar curtailments by the California ISO~\cite{CAISO_ManagingOversupply_2022}. 
    	Leveraging this renewable excess energy in FL can drastically reduce its operational carbon emissions.}
    \vspace{-2mm}
    \label{fig:caiso}
\end{figure}

The majority of today's machine learning (ML) solutions perform centralized learning, where all required training data are gathered in a single location, usually an energy-efficient data center with specialized hardware.
Yet, in many practical use cases, it is not feasible to collect data across a distributed system due to security and privacy concerns or because large amounts of raw data cannot be migrated from the deep edge to the cloud.
Federated Learning (FL) was introduced to address this issue by enabling distributed training of ML models without transmitting training data over the network~\cite{McMahan_FL_2016}.
In FL, we train a common ML model on clients that cannot or do not want to share their data, by iteratively distributing the model to a subset of them.
Clients then train locally on their own data and send back the updated models to the server, which aggregates them before starting the next round.

Unfortunately, FL approaches require considerably more training rounds than traditional ML and are often executed on infrastructure that is less energy-efficient than centralized GPU clusters, resulting in a significant increase in overall energy usage and associated emissions~\cite{Qiu_FirstLookCarbonFootprintFL_2021, Yang_EnergyEfficientFLWireless_2021, Naidu_CarbonEmissionsDifferentiallyPrivate_2021}.
Even without the application of FL, the training of large ML models is known to be an energy-hungry process and has increasingly raised concerns in recent years~\cite{Wu_SustainableAIMLSys_2022, Strubell_EnergyConsiderationsDLResearch_2020, Dodge_CarbonIntensityAICloudInstances_2022}.
As models keep growing in size and complexity, this problem is expected to aggravate, which is why there are numerous efforts towards more energy-efficient algorithms and hardware to reduce the carbon footprint of AI. %
Yet, when focusing on reducing emissions, ``using renewable energy grids for training neural networks is the single biggest change that can be made”~\cite{Payal_NatureCarbonImpactAI_2020, Google_CarbonFootprintMLShrink_2022}.

In this work, we study how the operational carbon emissions of synchronous FL training can be reduced to zero by operating under the hard constraint of only leveraging renewable excess energy and spare computing capacity at cloud or edge resources.
Excess energy, also called stranded energy, occurs in electric grids when more power is generated than demanded or when the grid does not have sufficient capacity for transmission.
If the oversupply cannot be stored in batteries (which are expensive and only available in a limited capacity) or traded with neighboring grids (whose excess energy patterns often correlate) the last resort is curtailment, the deliberate reduction in production.
Through curtailment, the California Independent System Operator wasted more than 27 million megawatt-hours of utility-scale solar energy in 2022, which is around 7\,\% of their entire solar production~\cite{CAISO_ManagingOversupply_2022}. 
Due to the increasing penetration of variable renewable energy sources, the amount of curtailed energy is only expected to grow, as shown in Figure~\ref{fig:caiso}.
At the same time, many existing computing infrastructures are frequently underutilized or could be overclocked if the occurrence of excess energy justifies reduced energy efficiency~\cite{Chien_BeyondPUE_2022}. 
To make better use of these resources, \emph{carbon-aware computing}, i.e. considering the spatio-temporal availability of low-carbon energy during scheduling, has attracted much attention in recent years~\cite{Radovanovic_Google_2021, Wiesner_LetsWaitAwhile_2021, Fridgen_NotAllDoomAndGloom_2021, Bashir_SustCompWithoutHotAir_2022, Lin_CouplingDatacentersPowerGrids_2021, Zheng_MitigatingCurtailment_2020, Wiesner_Cucumber_2022}.

FL is a promising workload for carbon-aware computing, as it consists of energy-intensive batch jobs that are scheduled in geo-distributed environments (to leverage spatial resource and energy availability) without strict runtime requirements (to also leverage temporal variations).
However, as excess energy and the availability of spare computing resources can be highly volatile, not explicitly taking them into account during client selection can lead to significantly longer training times due to \emph{stragglers}: clients that perform less local training than expected, or become entirely unavailable during a training round.
Furthermore, energy-agnostic selection strategies can introduce biases by disproportionately selecting clients that have a lot of excess resources available throughout the training.
Yet, the idea of aligning FL scheduling with the availability of renewable energy has so far only been studied theoretically and under assumptions like independent and identically distributed (iid) data and fixed ``energy arrival" patterns that are not realistic in practice and do not consider the above challenges~\cite{Guler_SustainableFederatedLearning_2021}.

To fill this gap, we propose \emph{FedZero}, an FL system for heterogeneous and geo-distributed environments that utilizes forecasts for renewable excess energy and spare computing capacity to ensure fast, efficient, and fair training under energy and resource constraints.
We summarize our \textbf{contributions} as follows:

\begin{itemize}
	\item We propose a system design for executing FL trainings exclusively on renewable excess energy and spare computing capacity which allows clients to share common energy budgets at runtime.
	\item We introduce a scalable client selection strategy that results in fast convergence and fair client participation under variable energy and resource constraints.
	\item We evaluated our approach on different datasets, models, and scenarios, to show that FedZero enables fast and energy-efficient training while being robust to forecast errors.
	\item We implemented FedZero and all baselines using Flow\-er~\cite{Beutel_Flower_2020} and Vessim~\cite{wiesner2023vessim} and made this code openly available\footnote{\url{https://github.com/dos-group/fedzero}}.
\end{itemize}

\section{A Case for FL on Excess Resources}
\label{sec:background}

FL was originally developed for use cases on mobile and edge devices, where individual clients usually consume little energy.
However, in recent years a variety of new application domains have been explored to enable cross-device and cross-silo training, many of which include clients with significant computing capabilities and electricity demand.
Examples of these novel FL settings include healthcare~\cite{Rieke_FLHealth_2020}, the financial sector~\cite{Yang_FLMorganClaypool_2019}, remote sensing~\cite{So_FedSpace_2022}, autonomous driving~\cite{Nguyen_FLAutonomousDriving_2022}, and smart cities~\cite{Jiang_FLSmartCity_2020}, which all consist of complex models that require periodic re-training to adapt to changing environments. 

We argue that in environments where individual clients require significant energy for participating in an FL training, it merits to explicitly consider the availability of renewable excess energy in client selection and during training to reduce carbon emissions.

\subsection{Renewable Excess Energy}
\label{sec:background_energy}

Due to the expanding deployment of variable renewable energy sources such as solar and wind, it is becoming increasingly challenging to match power supply and demand at all times.
If locally occurring renewable excess energy cannot be passed on to neighboring grids due to limited grid capacity and cannot be buffered in some kind of energy storage, the only option left to operators is to throttle supply.
In this section, we describe the two main scenarios in which renewable excess energy can occur.

The most direct way of operating IT infrastructure in a sustainable manner is through the use of on-site renewable energy, where the energy source is located close to the datacenter or powerful edge device.
Within a microgrid, energy storage can buffer limited amounts of excess energy but it is expensive, entails losses, and frequent charge cycles accelerate battery aging~\cite{Liu_BatteryAging_2017}.
Moreover, while more and more countries offer the possibility to sell energy to the public grid, feed-in tariffs are usually well below purchase prices~\cite{REN21_RenewablesReport_2022}. 
Therefore, operators have a clear incentive to consume all generated electricity directly.

The more common practice to achieve datacenters ``powered by 100\,\% renewable energy", as claimed by big cloud service provi\-ders~\cite{GoogleCarbonFree, MicrosoftCarbonFree, AmazonCarbonFree}, is through carbon accounting.
Carbon accounting allows operators to offset their consumption of carbon-intensive grid energy through the purchase of renewable energy certificates.
However, today's certificates allow that power production and consumption can take place at vastly different times and locations, which is why their utility for achieving science-based targets is often questioned~\cite{Bjorn_RECThreatenTargetsNature_2022}.
A prominent effort towards stricter carbon accounting, often called \emph{24/7 matching}, are Google's Time-based Energy Attribute Certificates (T-EACs)~\cite{Google_CarbonCounting_2021} that are issued hourly and location-specific~\cite{Bashir_SustCompWithoutHotAir_2022}.
Similar ideas have been brought forward by Microsoft~\cite{Amazon_CarbonCounting_2022} and Amazon~\cite{Microsoft_CarbonCounting_2021}.
During curtailment periods, these certificates are expected to be very cheap and can be used for the cost-effective execution of flexible workloads such as FL.

\subsection{Computing on Spare Resources}

As we do not want to promote the purchase of new hardware just to enable the flexible scheduling of training jobs, FedZero aims to schedule workloads exclusively on spare capacity of existing hardware.
We believe this is realistic in a wide range of scenarios as many IT infrastructures are over-dimensioned to accommodate peak loads.
For example, public clouds, as well as emerging public edge solutions, must maintain a relatively high proportion of spare computing resources to sell their promise of infinite scalability.
Similarly, on-site infrastructure is often designed for peak loads and can be severely underutilized outside these periods.
A common measure in cluster managers to increase the utilization during off-peak hours is by deferring delay-tolerant workloads, for example in the form of \emph{best-effort} jobs~\cite{Tirmazi_BorgNextGen_2020}.

Lately, several ideas have been developed to not only shift load for increased resource utilization but for better aligning electricity usage with grid carbon-intensity or excess energy~\cite{Radovanovic_Google_2021, Lin_CouplingDatacentersPowerGrids_2021, Wiesner_LetsWaitAwhile_2021}.
Moreover, even infrastructure that is already utilized close to capacity can be used to compute flexible workloads, if the use of otherwise curtailed excess energy justifies the reduced energy efficiency caused by overclocking and increased cooling~\cite{Chien_BeyondPUE_2022}.

\section{Problem Statement}
\label{sec:problem_statement}

Our goal in this paper is to train a federated learning model with no operational carbon emissions in an efficient and fair manner.
We aim to optimize for fast convergence and low overall energy usage in a setting, where clients are only allowed to train on excess resources.
We do so by cherry-picking clients that are likely to have access to renewable excess energy and spare computing capacity (or potential for overclocking) and by operating within these constraints at runtime.

\subsection{Challenges}

FL on excess resources poses a number of new challenges that have not yet been addressed by existing approaches.

\parag{Convergence speed and efficiency}
The availability of excess resources can be highly variable.
Thus, clients that have access to excess energy and spare computing resources during client selection can run out of resources over the duration of a training round.
This leads to an increased number of stragglers that can severely harm training performance.
A common way to alleviate the impact of stragglers in FL is to select more participants in each round than actually needed, but to only wait for a number of early responses before aggregating the results and starting a new round.
The extent of over-selection can be adapted to the environment but is usually around 30\,\% in the related work~\cite{Bonawitz_FlAtScale_2019, Lai_Oort_2021, PyramidFL_2022}.
While this makes FL training more robust, it has the disadvantage of wasting computing capacity and energy, since the work of some clients is discarded in each round.
Moreover, if multiple clients reside in the same power domain, over-selection can actively harm training progress as more clients share the same limited power source at runtime rather than only attributing power to the most useful clients.

\begin{figure}[h]
    \centering
       \includegraphics[width=1\columnwidth]{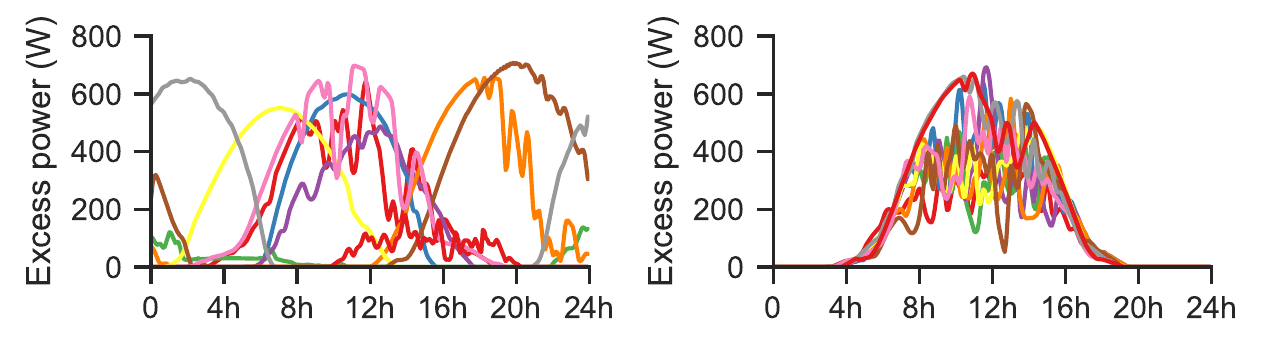}
       
    \vspace{-.5cm}\hspace{0\linewidth}\subfloat[
   		\label{fig:solar_global}
   		Distributed power domains.
   	]{\hspace{.47\linewidth}}
   	\hspace{.04\linewidth}
	\subfloat[
		\label{fig:solar_coloc}
		Co-located power domains.
	]{\hspace{.45\linewidth}}
    \vspace{-.3cm}\caption{Excess power availability for different scenarios.}
    \label{fig:power_domain}
    \vspace{-.3cm}
\end{figure}

\parag{Common power budgets}
As clients can share the same source of excess energy, we need to treat energy as a shared and limited resource during client selection and at runtime.
We use the term \emph{power domain} to describe the clustering of FL clients into groups with access to the same source of renewable excess energy -- either because they are physically connected within a datacenter's microgrid, or because their operation is covered through a common budget of, for example, T-EACs (see Section~\ref{sec:background_energy}).
Power domains are disjunct, meaning that one client can only be part of a single power domain.
Excess energy occurring within a power domain must be shared by all clients within the domain.
For example, in this work, we study the behavior of FL training under resource and energy constraints in two different solar-based scenarios:
In the first scenario, clients are spread across ten globally distributed power domains (Figure~\ref{fig:solar_global}).
In the second scenario, power domains are in close geographic proximity (Figure~\ref{fig:solar_coloc}).

\parag{Fairness of participation}
Simply optimizing for available resources in client selection will lead to a strong imbalance in favor of clients who have a lot of excess resources available throughout the training.
In realistic settings, where the distribution of data varies between clients, this can lead to significant biases toward certain data in the training, which is unfair and ultimately harms model performance.
Hence, even if the availability of excess resources is highly imbalanced, we want to ensure that all clients are able to participate in a similar number of rounds.

\parag{Robustness and scalability}
To target the previous challenges, we try to spread the client selection over different power domains using forecasts of available excess energy and spare computing resources -- while still considering system and statistical utility.
However, forecasts usually come with a certain error, which is why we require a solution that is robust to inaccurate forecasts.

Moreover, we need to ensure that any underlying optimization comes with a low overhead and runtime complexity, as real-life FL scenarios can comprise large numbers of clients.

\begin{table}[t]
\small
\centering
\caption{Overview of constants and variables}
\label{tab:overview}
\begin{tabular}{lp{5.5cm}}
\toprule
\multicolumn{2}{c}{System-related constants} \\
\midrule
$C$  & set of clients \\
$P$ & set of power domains  \\
$C_{p}$  & set of clients in power domain $p$ \\
$m_{c}$  & maximum capacity of client $c$ (batches/timestep) \\
$\delta_{c}$  & energy efficiency of client~$c$ (energy/batch) \\
\midrule
\multicolumn{2}{c}{User-defined constants} \\ 
\midrule
$n$  & number of selected clients per round \\ 
$d^\text{max}$  & maximum round duration in multiples of $t$ \\
$m_c^\text{min}; m_c^\text{max}$  & minimum/maximum number of batches client $c$ must participate per round \\
\midrule
\multicolumn{2}{c}{Input variables (updated each round)}
\\ \midrule
$m_{c,t}^\text{spare} \in [0,m_{c}]$  & spare capacity forecast for client~$c$ at time~$t$ \\ 
$r_{p,t}$  & excess energy forecast for power domain $p$ at time~$t$ \\
$\sigma_{c}$  & fairness weighting of client~$c$ \\ 
\midrule 
\multicolumn{2}{c}{Optimization variables (determined each round)} \\
\midrule
$d$ & expected round duration \\
$b_{c} \in \{0, 1\}$  & whether or not client $c$ is selected \\ 
$m_{c,t}^\text{exp} \in [0, m_{c,t}^\text{spare}]$  & expected number of batches client $c$ will compute at time $t$ considering energy and capacity constraints \\
\bottomrule
\end{tabular}
\end{table}

\subsection{Problem Formalization}

We define $C$ as the set of clients distributed over a disjunct set of power domains~$P$.
Clients are characterized by their energy efficiency~$\delta_c$ and maximum computing capacity~$m_c$.
For simplicity, we do not consider other resources like memory in this work, but they can be integrated the same way as $m_c$.
We divide time into slots of duration $t$, so an estimated training round duration $d$ is always a multiple of $t$.
The duration of $t$ depends on the problem setting but is usually in the order of one minute.
We define the training of a mini-batch, from now on called batch, as an atomic operation to be performed within these time slots.
Table~\ref{tab:overview} provides an overview of all introduced variables and constants.

As common for FL in heterogenous environments~\cite{Bonawitz_FlAtScale_2019, Li_FedProx_2020}, we allow clients to train a variable amount of batches, but require the configuration of a lower ($m_c^{min}$) and upper ($m_c^{max}$) bound per client (for example, 1 to 5 local epochs).
Furthermore, the server should define the number of selected clients per round $n$ as well as a maximum round duration $d^{max}$ after which results get aggregated, even if not all clients did respond in time.
We allow multiple clients to share a common excess energy budget by clustering them into power domains.
$C_p$ describes the clients of a power domain $p \in P$.
Each power domain comprises a control plane, like an ecovisor~\cite{Souza_Ecovisor_2023}, which is responsible for attributing power to clients.
Section~\ref{sec:client_selection} describes our client selection algorithm and optimization problem.

\subsection{Boundaries}
In this work, we do not explicitly consider energy storage or feeding excess energy to the public grid, since these options are not always available and have drawbacks compared to consuming excess energy directly~\cite{Wiesner_Cucumber_2022}.
Moreover, as we target larger-scale infrastructure that is usually connected to the network via highly energy-efficient optical fiber, we do not model the energy usage for data transmission.
Lastly, we require the availability of excess resources during training. Environments where relevant clients never have access to renewable excess energy or spare computing capacity need to default to a less radical approach and consider carbon-intensive grid energy consumption at times.

\section{System Design}
\label{sec:approach}

\begin{figure*}
    \centering
       \includegraphics[width=1\textwidth]{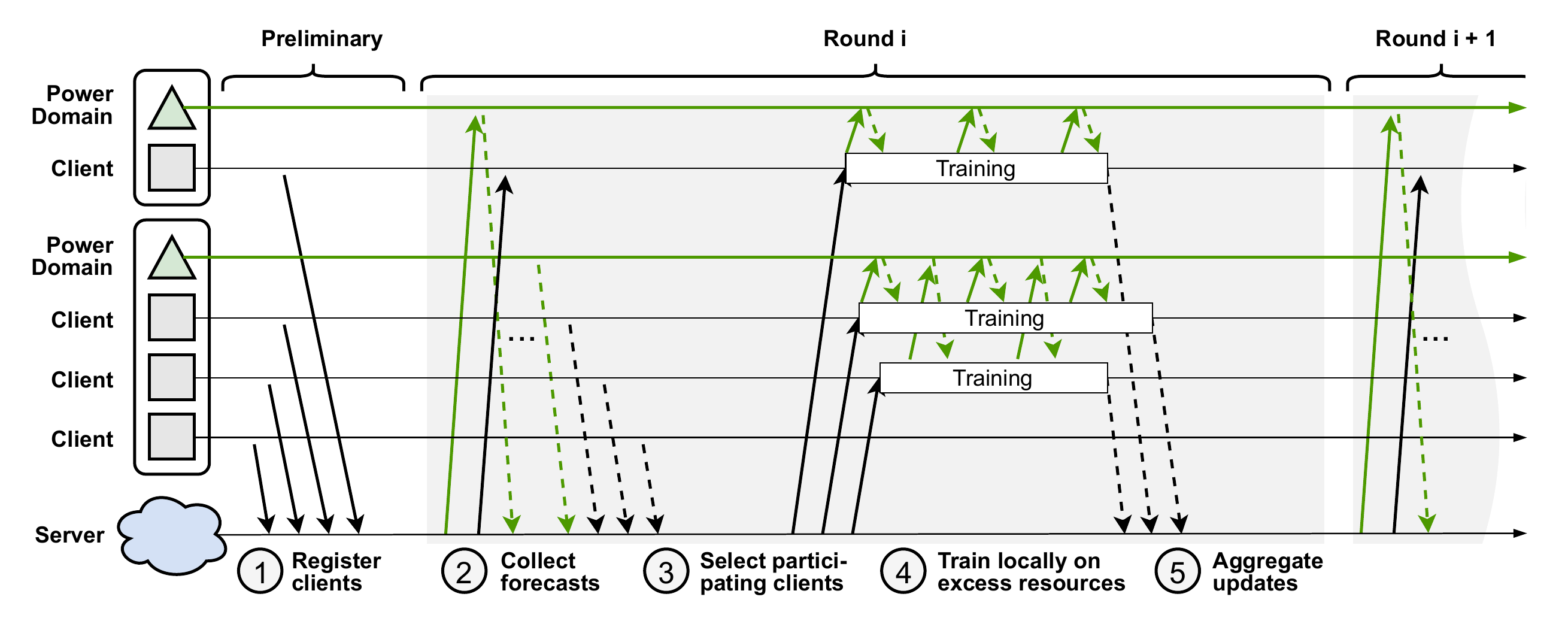}
    \caption{At each training round, FedZero queries excess energy forecasts of power domains and load forecasts of individual clients. Based on this information, it selects a set of clients for which it expects a short round duration at high statistical utility. At runtime, clients have to periodically adjust their training performance to align with the actual available excess energy.}
    \label{fig:overview}
\end{figure*}

An overview of FedZero's protocol is depicted in Figure~\ref{fig:overview}.
The training starts after a required amount of clients register themselves with the server (Section~\ref{sec:client_registration}, \circled{1}).
At the beginning of each round, the server requests forecasts on expected excess energy within power domains and spare capacity at clients  (Section~\ref{sec:local_forecasting}, \circled{2}).
FedZero then selects $n$ clients for training, for which it expects the shortest round duration under the given resource constraints (Section~\ref{sec:client_selection}, \circled{3}).
This selection is performed based on the forecasts and information on past participation or statistical utility of clients for ensuring performance and fairness (Section~\ref{sec:fairness}).
Next, the selected clients train locally on spare capacity and, via continuous exchange with the power domain controller, excess energy (Section~\ref{sec:local_training}, \circled{4}).
Finally, all participating clients send their updated model back to the server which aggregates them and documents the participated batches and local loss for future decisions (\circled{5}).

\subsection{Client Registration}
\label{sec:client_registration}

Before starting the training process, FedZero requires the following information for each client:

\begin{enumerate}
	\item The \emph{maximum computing capacity} of a client is denoted as $m_{c}$ (batches/timestep) and can be derived from its FLOPS (floating point operations per second), the model's MACs (multiply–accumulate operations), and the batch size. Alternatively, it can also be benchmarked before or during the training. For variable capacity datacenters~\cite{Chien_BeyondPUE_2022}, $m_c$ should describe the actual maximum with overclocking.
	\item The \emph{energy efficiency} is denoted as $\delta_{c}$ (energy/batch) and can be obtained through measurements or derived from the client's system performance and power consumption characteristics. Linear power modeling is a meaningful simplification if we can assume power-proportional clients or sequential processing of workloads. If not, $\delta_{c}$ can also change from round to round depending on the system utilization, which is especially relevant in variable-capacity datacenters~\cite{Chien_BeyondPUE_2022}.
	\item The \emph{control plane addresses} define a client’s power domain, as well as where to query load and excess energy forecasts from. Load forecasts can be provided by the client itself or its cluster manager/container orchestrator. Typical providers for energy forecasts are electricity providers, microgrid control systems, or ecovisors~\cite{Souza_Ecovisor_2023}.
\end{enumerate}

\subsection{Forecasting Excess Energy and Load}
\label{sec:local_forecasting}

To avoid picking clients with access to little or no resources during a round, FedZero relies on multistep-ahead forecasts of excess energy at power domains and spare capacity at clients.

Power production forecasts for variable renewable energy sources like solar~\cite{BRIGHT2018118satellitederived,Khalyasmaa2019PredictionOS} and wind~\cite{Alencar2017DifferentMFwindpowergenerationcasestudy,LI2021121075powerpredictionwindturbinesgaussian} are usually based on weather models for mid- and long-term predictions as well as, in case of solar, satellite data for short-term predictions that enable up to 5-minute resolution.
For on-site installations, there exist a large number of companies providing power production forecasts as a service.
In the case of time-based power purchase agreements, it is the responsibility of the utility provider to inform their customers of future energy budgets.
For determining future excess energy, the system furthermore needs to take load forecasts for IT infrastructure as well as co-located consumers into account.
We define $r_{p,t}$ to be the forecasted excess energy of power domain $p$ at time $t$.

Load prediction is a widely researched field covering forecasts related to application metrics, such as requests per second, as well as the utilization of (virtualized) hardware resources like CPU, GPU, or RAM.
They usually entail time series forecasting models trained on historical data but can also take additional context information into account.
As $m_c$ describes the maximum computing capacity of a client in batches/timestep, we define $m_{c,t}^\text{spare} \in [0,m_{c}]$ to be its forecasted spare capacity at time $t$.

\subsection{Client Selection}
\label{sec:client_selection}

\begin{algorithm}[b]
    \caption{Determine clients and round duration
    }
    \algrenewcommand\algorithmicindent{1.0em}
    \label{alg:1}
    \begin{algorithmic}[1] %
    \State $C \gets \text{set of clients}$
    \State $P \gets \text{set of power domains}$
    \LineComment{search for shortest possible round duration}
    \For{$d \gets 1$ \textbf{to} $d^\text{max}$}
    	\LineComment{filter out power domains without excess energy}
            \State $\bar{P} \gets \{ \forall p \in P, \forall t = 1,...,d : r_{p,t} > 0\}$ \label{algl:filter_p}
		
    	\LineComment{filter out clients that over-participated in the past (see Section~\ref{sec:fairness})}
		\State $\bar{C} \gets \{ \forall c \in C : \sigma_{c} > 0\}$\label{algl:filter_c2}
		
		\LineComment{filter out clients without sufficient computing capacity or energy}
		\For{$p \in \bar{P}$}
   			\State $\bar{C} \gets \bar{C} \setminus \{ \forall c \in C_p : \sum_{t = 0}^{d} \min(m_{c,t}^\text{spare}, \frac{r_{p,t}}{\delta_c}) < m_c^{min}\}$\label{algl:filter_c1}
		\EndFor
		
		\LineComment{increase duration if there are not at least $n$ valid clients}
		\If{$|\bar{C}| < n$} %
			\State \textbf{continue}
		\EndIf

    	\LineComment{select optimal clients}
		\State $b$ $\gets$ findOptimalClients$(\bar{C}, \bar{P}, d)$
		\If{$b$ is valid solution} \label{algl:mip}
			\State \textbf{return} $b, d$
		\EndIf
	\EndFor
	\LineComment{wait, if no solution is found for $d = d^{max}$}
    \end{algorithmic}
\end{algorithm}

FedZero selects clients based on their forecasted energy and capacity constraints as well as statistical utility.

\parag{Iterative search}
FedZero optimizes for system utility by selecting clients that are expected to compute their $m_c^{min}$ as fast as possible.
We guarantee low computational overhead, by performing an iterative search over possible round duration~$d$: For each round duration, we solve a simple mixed-integer program (MIP) which scales linearly with the number of clients and power domains (see Section~\ref{sec:eval_scalability}).
For simplicity, the iterative search is described as an incrementing for-loop in Algorithm~\ref{alg:1}. %
In practice, it can be implemented as a binary search with complexity $\mathcal{O}(\log{}n)$.

On every iteration, Algorithm~\ref{alg:1} heavily pre-filters entire power domains (Line~\ref{algl:filter_p}) and individual clients (Lines~\ref{algl:filter_c1} and~\ref{algl:filter_c2}) that cannot constitute valid solutions within the current $d$, to further reduce the runtime of the MIP.
If no valid solution is found within the maximum round duration $d^{max}$, the algorithm waits for conditions to improve or it cloud resolve the situation by weakening constraints, e.g. by considering grid energy.
In this work, we only operate under hard energy and capacity constraints.

\parag{Optimization Problem}
For the MIP, we define two discrete optimization variables per eligible client:
\begin{itemize}
	\item $b_{c} \in \{0,1\}$ equals 1 iff client $c$ participates in the round
	\item $m_{c,t}^\text{exp} \in [0, m_{c,t}^\text{spare}]$ denotes the expected number of batches client $c$ will compute at time $t$
\end{itemize}

\noindent
The optimization problem is described as follows:
\begin{align}
\max_{b_c,m_{c,t}^{\text{exp}}}\quad &\sum_{c \in C} b_{c} \cdot \sigma_{c} \sum_{t = 0}^{d} m_{c,t}^\text{exp}\nonumber \\ %
\text{s.t.}\quad &b_c = 1 \implies m_c^\text{min} \leq \sum_{t = 0}^{d} m_{c,t}^\text{exp} \leq m_c^\text{max} \quad \forall c \in C \label{eq:const_exp}\\
&\sum_{c \in C_{e}} m_{c,t}^\text{exp} \cdot \delta_{c} \leq r_{e,t} \quad \quad \quad \,\forall e \in E,t = 0, \ldots, d \label{eq:const_energy}\\
&\sum_{c \in C} b_{c} = n \label{eq:const_selection}
\end{align}

We optimize for the maximum number of batches to be computed within the input duration $d$, weighted by each client's statistical utility $\sigma_{c}$. 
Equation~(\ref{eq:const_exp}) limits each selected client to compute between $m_c^{min}$ and $m_c^{max}$ batches.
Equation~(\ref{eq:const_energy}) constrains all clients in a power domain to not use more energy than available.
Equation~(\ref{eq:const_selection}) ensures that exactly $n$ clients are selected per round.

\parag{Statistical utility}
We introduce a utility function 
$f: C \to \{\sigma_{c} : \forall c \in C\}$
which is invoked in every round and returns a weighting that gives precedence to certain clients in the optimization problem.
This function can be based on the previous participation of clients, an approximation of statistical client utility, or other user-defined metrics, for example, to respect fairness constraints like group parity.

The utility function applied in the remainder of this paper is based on the statistical utility function proposed in Oort~\cite{Lai_Oort_2021}:
$$
\sigma_{c} =
\begin{cases}
    |B_c|\sqrt{\frac{1}{|B_c|} \sum_{k \in B_c} loss(k)^2}, & \text{if } p(c) \geq 1\\
    1, & \text{otherwise}
\end{cases}
$$
Oort approximates statistical client utility based on the number of available training samples $B_c$ and the local training loss, which is expected to correlate with the gradient norm.

\subsection{Ensuring Fair Participation}
\label{sec:fairness}

When performing FL in heterogenous environments under excess energy and capacity constraints, we actively have to take care of avoiding biases towards powerful clients with lots of spare capacity ($m_{c}$) or clients within power domains with large amounts of excess energy ($r_{e}$).
This problem is exacerbated by FedZero, which prefers energy-efficient clients ($\delta_{c}$) and -- without further measures -- tends to select similar sets of clients in consecutive rounds, which can harm the model's generalization performance.

To mitigate the mentioned biases and reduce variance, we add clients to a blocklist after they participate in a training round.
Blocked clients get assigned $\sigma_c = 0$ and are hence excluded from future rounds.
At the start of each round, clients can get released from the blocklist with probability $P(c)$:
$$
P(c) =
\begin{cases}
    (p(c) - \omega)^{-\alpha}, & \text{if } p(c) - \omega > 0\\
    1, & \text{otherwise}
\end{cases}
$$

\noindent
where $p(c)$ describes the number of rounds a client previously participated and $\alpha$ is a user-defined parameter that controls the speed at which clients get released.
A high $\alpha$ will cause overparticipating clients to remain longer on the blocklist, thereby reducing the set of clients that FedZero can pick from.
This can extend training time but ensures fair participation.
An $\alpha$ close to 0 reduces the impact of the blocklist.
We consider $\alpha=1$ for the remainder of this paper, which turned out to provide the best balance between training speed and performance in all evaluated experiments.

The parameter $\omega$ avoids decreasing release probabilities over time and gets periodically updated to $\omega = \text{mean}\{p(c) : \forall c \in C\}$.
Users can choose a different $P(c)$ for their use case, for example, to improve group fairness or other custom metrics.

\subsection{Executing Training Rounds}
\label{sec:local_training}

This section describes the local control loop executed by clients during training rounds, see \circled{4} in Figure~\ref{fig:overview}.
Using the actually available resources at runtime, each client tries to compute $m_c^{min}$ batches as fast as possible.
Upon completion, it notifies the server but continues computation until $m_c^{max}$ is reached.
The server signals the end of a training round and gathers all updated models once all clients computed $m_c^{min}$, or once $d^{max}$ has passed.
If a client does not manage to compute at least $m_c^{min}$ batches before $d^{max}$, its work is discarded to not impede the training progress, as commonly performed in the literature~\cite{Bonawitz_FlAtScale_2019}.

Below, we discuss the two main challenges of the local control loop:
First, if multiple clients from the same power domain are participating in the same round, they have to share a common energy budget at runtime.
Second, the actual available excess energy and spare capacity are subject to short-term fluctuations and usually differ from previously performed forecasts.

\parag{Sharing power at runtime}
If only one client within a power domain is participating, it can make use of all available excess energy at runtime.
In this case, the capacity available for training is defined as the minimum of the free capacity and the capacity that can be powered using excess energy~\cite{Wiesner_Cucumber_2022}.

However, if two or more clients of a power domain participate simultaneously and there is not enough energy for all of them, they have to share a common energy budget at runtime, which has to be coordinated by the power domain controller.
To determine each client's share of power, we propose a simple two-step approach: 
First, power is attributed to clients that have not yet reached their minimum round participation $m_c^{min}$, weighted by how much energy is still required to reach the threshold.
If $m_c^{comp}$ describes the number of batches a client has already computed in the active round, this can be written as:
$C_p \to \{min(0,\ \delta_c m_c^{min} - \delta_c m_c^{comp}) : \forall c \in C_p\}$.
Second, if there is still power left, it is attributed to all clients below their maximum participation $m_c^{max}$, again weighted by the energy required to reach this limit:
$C_p \to \{min(0,\ \delta_c m_c^{max} - \delta_c m_c^{comp}) : \forall c \in C_p\}$.
As clients oblige capacity constraints and may not be able to make use of their entire share of power, the actual distribution of power must be decided in constant consultation with clients.

\parag{Short-term variations}
As FedZero aims to not interfere with other processes running on a  client, the extent of local training must be adapted over time.
For example, to determine excess energy, clients must periodically query their power domain controller, e.g. ecovisor~\cite{Souza_Ecovisor_2023} and use this information to perform power capping of tasks or entire containers~\cite{Li_Thunderbolt_2020, Enes_PowerBudgetingCluster_2020}.
Recently, works have been proposed that do this in a more sophisticated manner than simply throttling the training process' access to resources.
For example, DISTREAL~\cite{Rapp_DISTREAL_2022} handles the time-varying availability of resources in FL by dynamically adjusting the computational complexity of the trained neural network.
This approach could be extended to also consider power consumption, which is ultimately composed of the utilization of computing resources.
However, runtime behavior in the presence of short-term resource and energy fluctuations is currently not considered in our prototype.

\section{Evaluation}
\label{sec:evaluation}

We implemented FedZero and all baselines using the FL framework Flower~\cite{Beutel_Flower_2020} and the energy system simulator Vessim~\cite{wiesner2023vessim}.
We extended Flower to enable discrete-event simulation over time series datasets like excess energy availability and client load.
It enables us to perform experiments faster than in real-time by, for example, skipping over time windows where the system is idle and waiting for excess energy or spare capacity on clients.
We use Gurobi\footnote{\url{https://www.gurobi.com}} to solve the MIP.
Our implementation and all datasets are openly available (see Section~\ref{sec:intro}).

\subsection{Experimental Setup}
\label{sec:experimental-setup}

To evaluate FedZero, we simulate the power usage characteristics and performance of 100 FL clients using our Flower extension and perform the training on four NVIDIA V100 and two RTX 5000 GPUs.
This allows us to evaluate our approach without training models over multiple weeks and consuming megawatt hours of energy.

\begin{table}[h]
\centering
\small
\caption{Max energy consumption and training performance of the three types of clients.}
\vspace{-2mm}
\label{tab:client-types}
\begin{tabular}{@{}cccccc@{}}
\toprule
client & max    & \multicolumn{4}{c}{performance (samples per minute)}  \\ \cmidrule(lr){3-6}
type & energy & DenseNet-121  & EfficientNet-B1 &  LSTM  & KWT-1         \\ \midrule
\textbf{small} & 70 W  &        110 &  118        &   276     & 87       \\
\textbf{mid}   & 300 W &        384 &  411        &   956     & 303      \\
\textbf{large} & 700 W &        742 &  795        &   1856    & 586      \\ \bottomrule
\end{tabular}
\end{table}

\parag{Clients}
We model heterogeneity among clients by randomly assigning them to one of three types (\emph{small}, \emph{medium}, \emph{large}) that are roughly based on the performance\footnote{\url{https://developer.nvidia.com/deep-learning-performance-training-inference}} and energy usage characteristics of T4, V100, and A100 GPUs, respectively.
However, we downscaled their actual compute capabilities (samples per minute), as shown in Table~\ref{tab:client-types}.
We use 100 randomly selected machines from the Alibaba GPU cluster trace dataset~\cite{Weng_AlibabaTraces_2022} to model client load (\texttt{gpu\_wrk\_util}) and load forecasts (\texttt{gpu\_plan}).

\begin{figure}
    \centering
       \includegraphics[width=1\columnwidth]{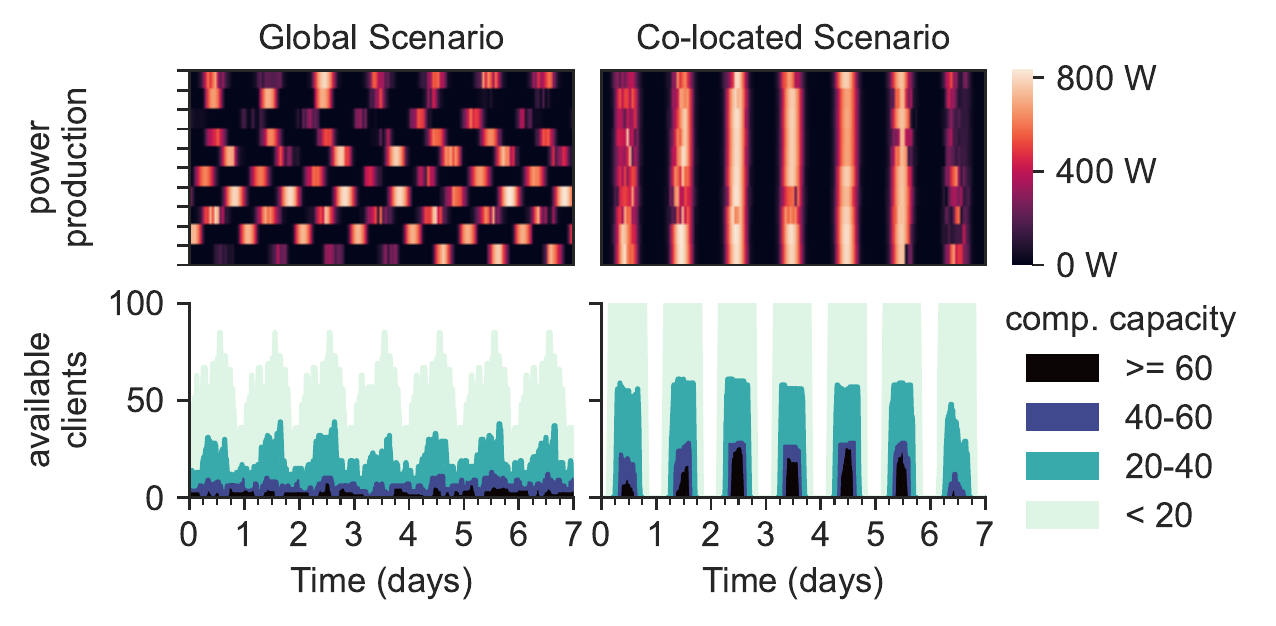}
    \vspace{-7mm}
    \caption{Power production and client availability over the course of both scenarios. While there are always some clients available in the global scenario, in the co-located scenario clients are always available around the same time.}
    \label{fig:power_and_availability}
    \vspace{-2mm}
\end{figure}

\parag{Scenarios}
For modeling power domains, we focus on on-site solar energy generation in two scenarios based on real solar and solar forecast data provided by Solcast\footnote{\url{https://solcast.com}}: A
\emph{global scenario} (ten globally distributed cities from June 8-15, 2022) and a \emph{co-located scenario} (ten largest cities in Germany from July 15-22, 2022), both displayed in Figure~\ref{fig:power_domain}.
The solar data is available in 5-minute resolution and we assume a constant power supply for steps within this period.
Clients are randomly distributed over the ten power domains, which each have a maximum output of 800\,W.
If there is little sun, or multiple clients are selected within a domain, energy becomes a limiting resource.
The power and client availability is depicted in Figure~\ref{fig:power_and_availability}.
The upper plot shows the energy availability within energy domains, where each domain is represented by one line.
The lower plot depicts the availability of clients over time, color-coded by how much of their total computational capacity is available for training.

\parag{Datasets, models, parameters}
We evaluate our approach on four datasets and models commonly used in FL evaluations.

\begin{itemize}
	\item \emph{CIFAR-100}~\cite{Krizhevsky09learningmultiple} contains 60,000 32x32 color images across 100 classes. 
We model heterogeneous data by applying a Dirichlet distribution with $\alpha = 0.5$, similar to~\cite{noniid_dirichlet}, which skews the number of samples as well as the number of samples per class and client.
We train\footnote{SDG optimizer , learning rate = 0.001, weight decay = 5e-4, momentum = 0.8} the convolutional model DenseNet-121~\cite{DenseNet_2017} using FedProx~\cite{Li_FedProx_2020} with $\mu = 0.1$.
	\item \emph{Tiny ImageNet} contains 100,000 64×64 color images of 200 classes.
We distribute samples to clients using the same Dirichlet distribution as for CIFAR-100.
We train\footnote{Adam optimizer, learning rate = 0.001} an Ef\-fi\-cient\-Net-B1~\cite{pmlr-v97-tan19a} model using FedProx with $\mu = 0.1$.
	\item In the \emph{Sheakespare}~\cite{Caldas_LEAF_2019} dataset, each client represents one of 100 randomly selected speaking roles from a play.
As in~\cite{Li_FedProx_2020}, we train a two-layer LSTM\footnote{100 hidden units, 8D embedding layer, SDG optimizer, learning rate = 0.8, see~\cite{Li_FedProx_2020}} using FedProx with $\mu = 0.001$ to perform next character prediction.
	\item \emph{Google Speech Commands} contains more than 100,000 audio samples of 30 different words. We randomly assigned speakers to the 100 clients and train\footnote{AdamW optimizer, learning rate = 0.001, weight decay = 0.1, see~\cite{berg21_interspeech}} the keyword transformer model KWT-1~\cite{berg21_interspeech} for speech classification.
\end{itemize}

\begin{table*}
  \centering
  \caption{Best accuracy and time/energy to reach the target accuracy of FedZero and the best-performing baselines.}
  \vspace{-3mm}
  \begin{tabular}{ccccccccccc}
    \toprule
    \multirow{3}[4]{*}{Dataset \& model} & \multirow{3}[4]{*}{Approach} & \multicolumn{4}{c}{Global scenario} &    & \multicolumn{4}{c}{Co-located scenario} \\
\cmidrule{3-6}\cmidrule{8-11}       &    & Target & Best & Time-to- & Energy-to- &    & Target & Best & Time-to- & Energy-to- \\
       &    & accuracy & accuracy & accuracy & accuracy &    & accuracy & accuracy & accuracy & accuracy \\
    \midrule
       & Random 1.3K & \multicolumn{1}{c}{\multirow{4}[2]{*}{64.7 \%}} & 66.0 \% & 4.7 d & 79.2 kWh &    & \multirow{4}[2]{*}{65.5 \%} & 66.6 \% & 5.3 d & 113.4 kWh \\
    CIFAR-100 & Oort 1.3K &    & 66.4 \% & 4.5 d & 103.8 kWh &    &    & 66.4 \% & 5.4 d & 138.7 kWh \\
    DenseNet-121 & Oort fc &    & 65.8 \% & 5.3 d & 102.4 kWh &    &    & 66.1 \% & 6.4 d & 126.7 kWh \\
       & \textbf{FedZero} &    & \textbf{66.8 \%} & \textbf{3.6 d} & \textbf{70.6 kWh} &    &    & \textbf{66.5 \%} & \textbf{4.5 d} & \textbf{96.4 kWh} \\
    \midrule
       & Random 1.3K & \multirow{4}[2]{*}{62.4 \%} & 63.1 \% & 5.6 d & 109.6 kWh &    & \multirow{4}[2]{*}{62.8 \%} & 63.3 \% & 3.7 d & 86.0 kWh \\
    Tiny ImageNet & Oort 1.3K &    & 63.2 \% & 3.3 d & 90.2 kWh &    &    & 63.5 \% & 3.4 d & 90.5 kWh \\
    EfficientNet-B1 & Oort fc &    & 63.1 \% & 3.9 d & 89.0 kWh &    &    & 62.7 \% & -  & - \\
       & \textbf{FedZero} &    & \textbf{63.6 \%} & \textbf{2.9 d} & \textbf{67.1 kWh} &    &    & \textbf{63.6 \%} & \textbf{3.4 d} & \textbf{75.8 kWh} \\
    \midrule
       & Random 1.3K & \multirow{4}[2]{*}{50.4 \%} & 50.7 \% & 4.6 d & 97.9 kWh &    & \multirow{4}[2]{*}{50.9 \%} & 51.5 \% & 4.5 d & 90.0 kWh \\
    Shakespeare & Oort 1.3K &    & 50.2 \% & -  & -  &    &    & 51.7 \% & 4.5 d & 95.4 kWh \\
    LSTM & Oort fc &    & 50.5 \% & 6.7 d & 157.4 kWh &    &    & 50.5 \% & -  & - \\
       & \textbf{FedZero} &    & \textbf{53.1 \%} & \textbf{1.8 d} & \textbf{40.0 kWh} &    &    & \textbf{53.1 \%} & \textbf{2.3 d} & \textbf{42.8 kWh} \\
    \midrule
       & Random 1.3K & \multirow{4}[2]{*}{83.6 \%} & 85.2 \% & 4.8 d & 103.5 kWh &    & \multirow{4}[2]{*}{82.8 \%} & 85.1 \% & 4.3 d & 80.8 kWh \\
    Google Speech & Oort 1.3K &    & 86.9 \% & 3.6 d & 99.0 kWh &    &    & 86.4 \% & 3.4 d & 85.0 kWh \\
    KWT-1 & Oort fc &    & 87.0 \% & 3.7 d & 86.2 kWh &    &    & 84.9 \% & 3.7 d & 76.6 kWh \\
       & \textbf{FedZero} &    & \textbf{87.2 \%} & \textbf{3.6 d} & \textbf{79.0 kWh} &    &    & \textbf{87.7 \%} & \textbf{2.6 d} & \textbf{65.8 kWh} \\
    \bottomrule
    \end{tabular}
    \label{tab:eval_results}
\end{table*}

The number of samples computed per timestep was obtained through benchmarking runs and is stated in Table~\ref{tab:client-types}.
All simulations use a timestep $t = 1$\, min and a max round duration $d^\text{max} = 60$\,min.
We select $n = 10$ clients each round which have to compute 1 to 5 local epochs, so $m^{min}$ and $m^{max}$ depend on the locally available number of samples.
Clients locally train on minibatches of size 10.
We ran each experiment five times over the course of seven simulated days and report mean values.

\parag{Baselines}
We compare FedZero with existing approaches by training six different baselines.
First, we run all experiments using \emph{Random} client selection as well as the guided selection strategy \emph{Oort}~\cite{Lai_Oort_2021}.
We update each client's system utility, an important factor in Oort's scheduling, based on the available energy and capacity in every round.
Both approaches can only select from clients, which currently have access to excess energy and spare resources.

Second, we train the above baselines again but this time allow them to select $1.3 n$ clients per round.
Over-selection is commonly employed in the related work~\cite{Bonawitz_FlAtScale_2019, Lai_Oort_2021} to counteract inefficiencies caused by stragglers in unreliable environments.
Once $n$ clients have returned their results a new round starts.
The baselines are called \emph{Random 1.3n} and \emph{Oort 1.3n}.

Third, we want to demonstrate that access to forecasts alone is not the decisive advantage over existing approaches.
For this, we train two baselines \emph{Random fc} and \emph{Oort fc} that only select 10 clients, but have access to load and energy forecasts for filtering out clients that are not expected to reach their $m_c^\text{min}$ within $d^{max}$.

Lastly, for each experiment, we define an \emph{Upper bound} in convergence speed and performance, by training a model that uses random client selection but is not subject to any energy constraints or existing load on clients (clients are still heterogeneous).
This baseline is not limited to renewable excess energy.

\subsection{Performance Overview}
\label{sec:eval_convergence_speed}

\begin{figure}[b]
    \vspace{-3mm}
    \centering
       \includegraphics[width=.99\columnwidth]{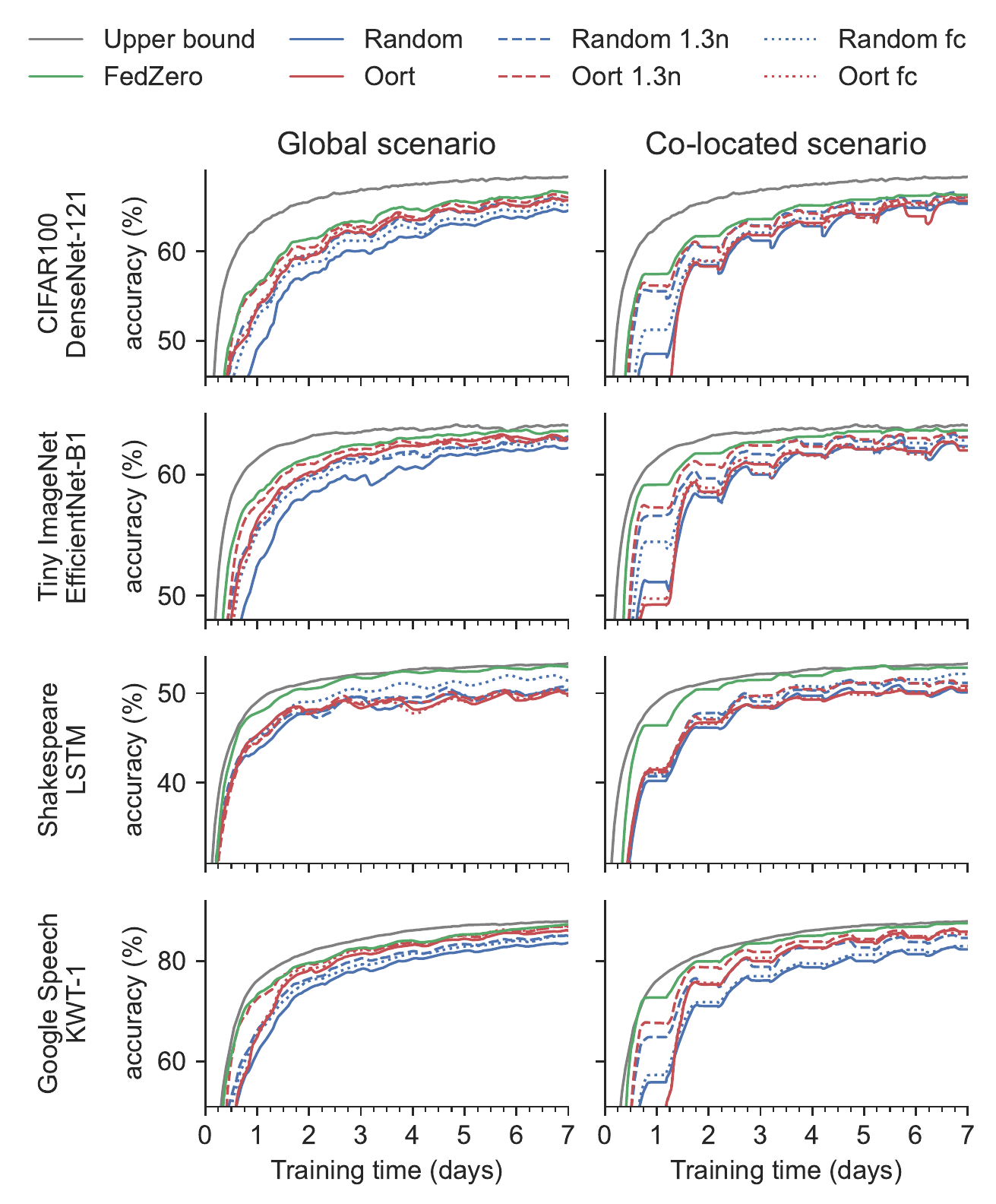}
    \vspace{-4mm}
    \caption{Training progress of all experiments.}
    \label{fig:training_progress}
\end{figure}

\parag{Training performance}
Figure~\ref{fig:training_progress} displays the training progress of FedZero and the baselines over the different experiments.
We can observe that FedZero consistently outperforms all baselines in terms of top accuracy.
While in some cases the performance of \emph{Oort}/\emph{Oort 1.3n}/\emph{Oort fc} is comparable (see Appendix~\ref{sec:appendix} for details), the gap is considerable in scenarios with heavy sample imbalance like Shakespeare ($2365\pm4674$ samples per client; $\text{min}=730$; $\text{max}=27950$).
This is due to the fact that none of the baselines considers common power budgets during client selection.
We found that this problem is exacerbated for strategies that select clients based on statistical utility, such as Oort:
If a power domain does not have access to excess energy for an extended period of time, the statistical utility of its clients is usually high as they have not participated for many rounds.
Once excess energy is available, Oort heavily targets these clients which leads to increased competition for energy at runtime and, therefore, slower training progress.

For all datasets but CIFAR-100, FedZero reaches, or almost reaches, the top accuracy of the \emph{Upper bound}, suggesting that varying resource availability does not necessarily harm training performance, but only increases the total training time.

\parag{Time-to-accuracy and energy-to-accuracy}
To further demonstrate how FedZero improves FL under energy and capacity constraints, we define the top accuracy of the \emph{Random} baseline as our target accuracy for a specific experiment. 
Table~\ref{tab:eval_results} reports the time and energy required for FedZero and the baselines to reach this target accuracy.
The table only contains the best-performing baselines; a full table with all results can be found in Appendix~\ref{sec:appendix}.

FedZero has the lowest time-to-accuracy and energy-to-accuracy across all experiments.
On average, it reached the target accuracy around 35\,\% faster in the global scenario and around 26\,\% faster in the co-located scenario than \emph{Random 1.3n} and \emph{Oort 1.3n}, which were among the fastest and most energy-efficient baselines.
At the same time, FedZero was using 36\,\% less energy on average in the global scenario and 30\,\% less in the co-located scenario.
Oort-based baselines generally outperform the Random-based ones in terms of top accuracy and convergence speed but at the cost of higher energy usage.
However, some Oort-based baselines do not reach the target accuracy at all, due to the previously described inefficiencies caused by selecting clients from the same power domain.

\parag{Round durations}
As FedZero knows about the system utility and resource availability, it avoids combining clients with vastly different expected round durations.
For example, in the global scenario on CIFAR-100 it required 15.1$\pm$8.5 min per round.
For comparison, the \emph{Random} baseline had an average round duration of 33.7$\pm$19.6 min, which was lowered to 22.7$\pm$17.7 min and 27.8$\pm$17.3 min by \emph{Random 1.3n} and \emph{Random fc}, respectively.
The round duration of Oort-based baselines was 18.6$\pm$14.5 min on average.

The same applies to the co-located scenario, where all clients are available around the same time.
Here, Random-based baselines take 15.5$\pm$12.7 min and Oort-based baselines 12.0$\pm$13.0 min on average.
FedZero only requires 9.7$\pm$7.6 min, allowing it to perform considerably more training rounds within the same time.
This observation is consistent across all experiments.

\subsection{Fairness of Participation}
\label{sec:eval_fairness}
\begin{figure}[t]
    \centering
    \includegraphics[width=1\columnwidth]{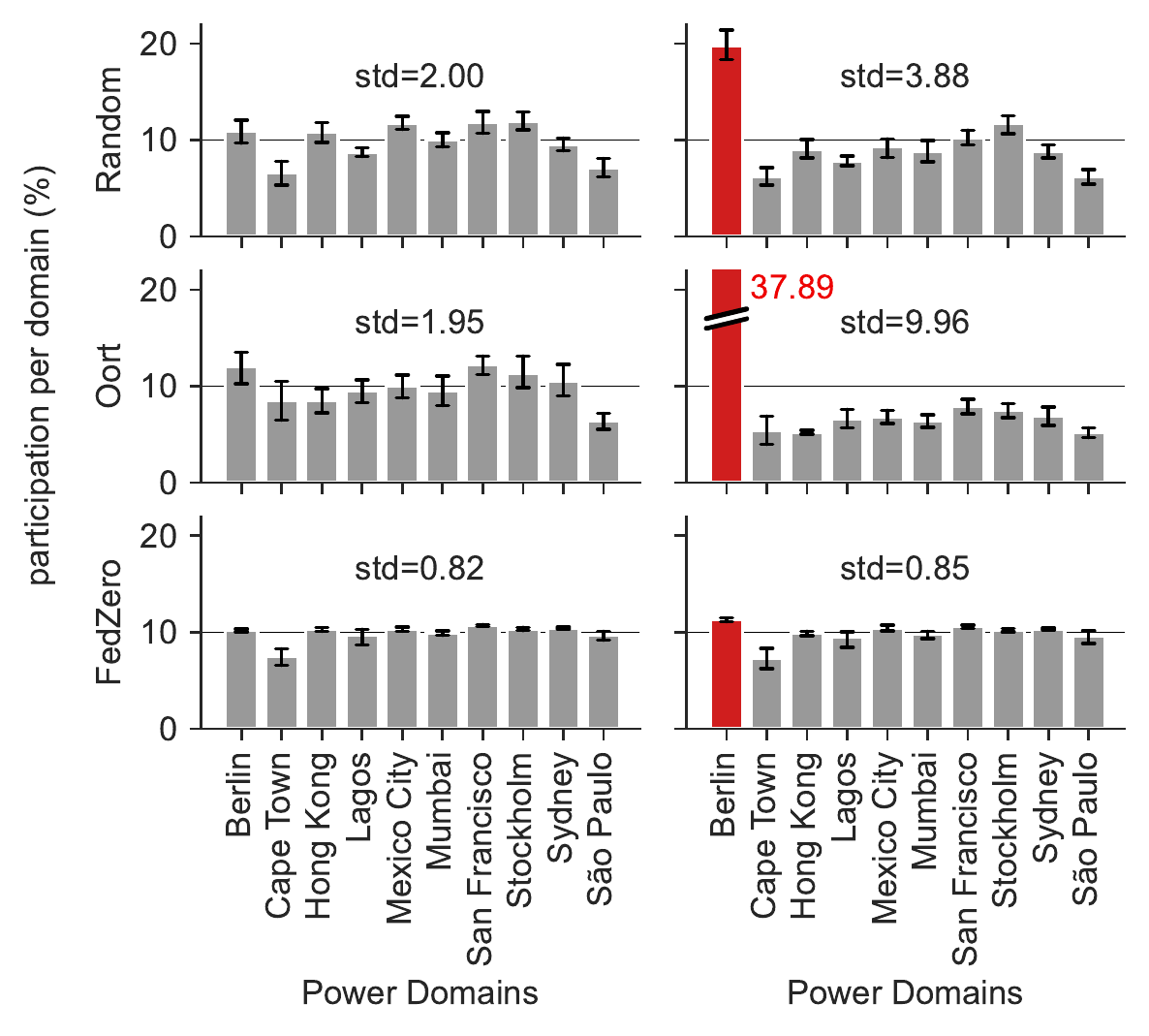}
    
   	\vspace{-0.5cm}\hspace{0.15\linewidth}\subfloat[
   		\label{fig:eval_fairness-a}
   		Client participation per power domain.
   	]{\hspace{.35\linewidth}}
   	\hspace{.03\linewidth}
	\subfloat[
		\label{fig:eval_fairness-b}
		Client participation with unlimited resources for \emph{Berlin}.
	]{\hspace{.45\linewidth}}
    \vspace{-1mm}\caption{FedZero ensures fair participation of clients, even under highly imbalanced conditions.}
\label{fig:eval_fairness}
\end{figure}

When training under energy and capacity constraints, we inevitably introduce biases towards clients that have lots of excess resources available.
To illustrate this, Figure~\ref{fig:eval_fairness-a} displays the average percentage of rounds in which clients have participated in the training for the CIFAR-100 global scenario, grouped by power domain.
As we select 10 out of 100 clients per round, we ideally expect an average client participation of 10\,\%.
However, some power domains have access to more excess energy than others, resulting in the \emph{Random} and \emph{Oort} strategies to favor clients in these domains.
We can observe that FedZero exhibits a much more balanced participation within (marked by the error bar) as well as between power domains (\emph{std} as stated on each figure).

\begin{table}[b]
  \centering
  \caption{CIFAR-100 performance on the global scenario under imbalanced conditions (Berlin has unlimited resources).}
  \vspace{-0.2cm}
    \begin{tabular}{cccc}
    \toprule
       & Best accuracy & Time-to-acc. & Energy-to-acc. \\
    \midrule
    Random & 64.6 \% & 6.7 d & 95.7 kWh \\
    Oort & 65.6 \% & 4.5 d & 189.4 kWh \\
    \textbf{FedZero} & \textbf{66.9 \%} & \textbf{3.5 d} & \textbf{83.4 kWh} \\
    \bottomrule
    \end{tabular}%
  \label{table:imbalanced}%
\end{table}%

We conducted an additional set of experiments on the same scenario, where the Berlin power domain has access to unlimited excess energy and all clients within Berlin have unlimited computing resources available.
The results of this experiment are displayed in Figure~\ref{fig:eval_fairness-b}, where Berlin is colored in red.
As clients in this power domain are now always available for training, the Random baseline almost doubles their participation from 11.0+-2.1 to 19.8+-2.6\,\%.
Even worse, Oort, which actively targets clients with high system utility, more than triples the participation of Berlin clients from 12.0$\pm$2.9 to 37.9$\pm$1.3\,\%.
Oort describes a mechanism for combining its selection with a user-defined fairness metric.
However, we found that Oort must rely almost entirely on our fairness metric, hence, disregard system and statistical utility, when attempting to achieve fairness comparable to FedZero.
Other than all baselines, which introduce significant biases, FedZero only slightly leverages the additional resources by increasing the mean participation of clients in the domain from 10.2$\pm$0.3 to 11.3$\pm$0.3\,\%.

Table~\ref{table:imbalanced} displays the training performance of the three approaches in the scenario where Berlin has unlimited resources.
We observe that \emph{Random} used 19\,\% more energy and \emph{Oort} even twice the energy in the imbalanced scenario (Figure~\ref{fig:eval_fairness-b}) for reaching a comparable accuracy as in the base scenario (Figure~\ref{fig:eval_fairness-a}).
FedZero used only 4\,\% more energy and reduced its time-to-accuracy.

Based on our results on fairness of participation, we expect FedZero to also improve other fairness metrics such as accuracy parity between clients.
However, these metrics depend on various additional factors like non-iid data distribution among power domains, which is why we leave this analysis to future work.

\subsection{Robustness Against Forecasting Errors}
\label{sec:eval_robust}

\begin{figure}[t]
    \centering
       \includegraphics[width=1\columnwidth]{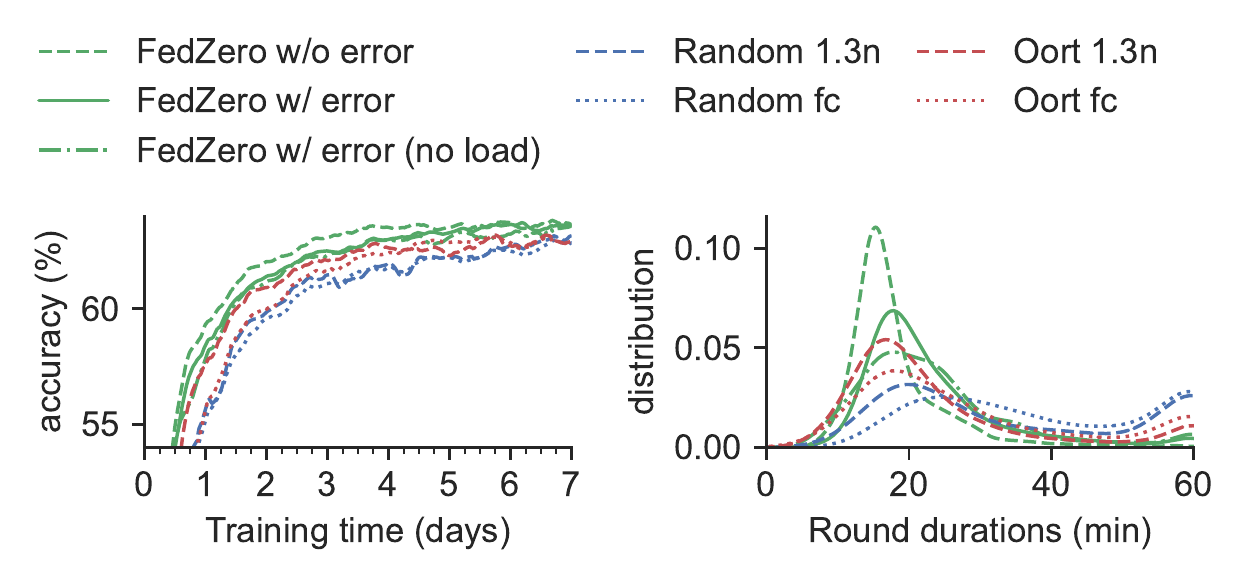}
    \vspace{-6mm}
    \caption{Analysis of convergence behavior and round durations of FedZero under forecasts of different quality.}
    \label{fig:fc_error}
\end{figure}

To investigate the impact of forecast quality on FedZero's performance, we performed further experiments based on the global scenario on the Tiny ImageNet and Google Speech datasets.
Figure~\ref{fig:fc_error} shows the training progress and distribution of round durations for Tiny ImageNet.
\emph{FedZero\;w/\;error} uses forecasts with realistic errors as in all previous experiments, \emph{FedZero\;w/o\;error} uses perfect forecasts, and \emph{FedZero\;w/\;error (no load)} uses realistic errors for excess energy but has no forecasts for spare capacity available, as short term load might not always be predictable in every setting.
Note, that FedZero is not able to operate if there are no predictions of excess energy at all, for example, due to communication loss.

The three experiments based on FedZero show small differences in convergence speed and energy usage.
While \emph{FedZero w/ error} takes 2.8\,d and 65.2\,kWh to reach the target accuracy, using perfect forecasts it requires 15.4\,\% less time and 15.2\,\% less energy.
This is because FedZero becomes better at avoiding stragglers through the use of accurate predictions, resulting in shorter, hence, more efficient rounds, as shown in the right of Fig~\ref{fig:fc_error}.
FedZero without load forecasts takes 8.2\,\% more time to reach the target accuracy, using 10.0\,\% more energy.
This result is of course specific to how we modeled load in our evaluation -- the effect in other contexts may be bigger or smaller.
Still, we can see that all three experiments converge to the same accuracy of 63.8\.\%, while consistently exhibiting better time-to-accuracy and energy-to-accuracy, showing that FedZero can perform well even with suboptimal forecast quality.

We additionally performed this analysis on the Google Speech experiment and got comparable results. For example, FedZero without errors converged 5.2\,\% faster using 6.7\,\% less energy than FedZero with realistic forecasts.

\subsection{Overhead and Scalability}
\label{sec:eval_scalability}

\begin{figure}[h]
    \centering
       \includegraphics[width=1\columnwidth]{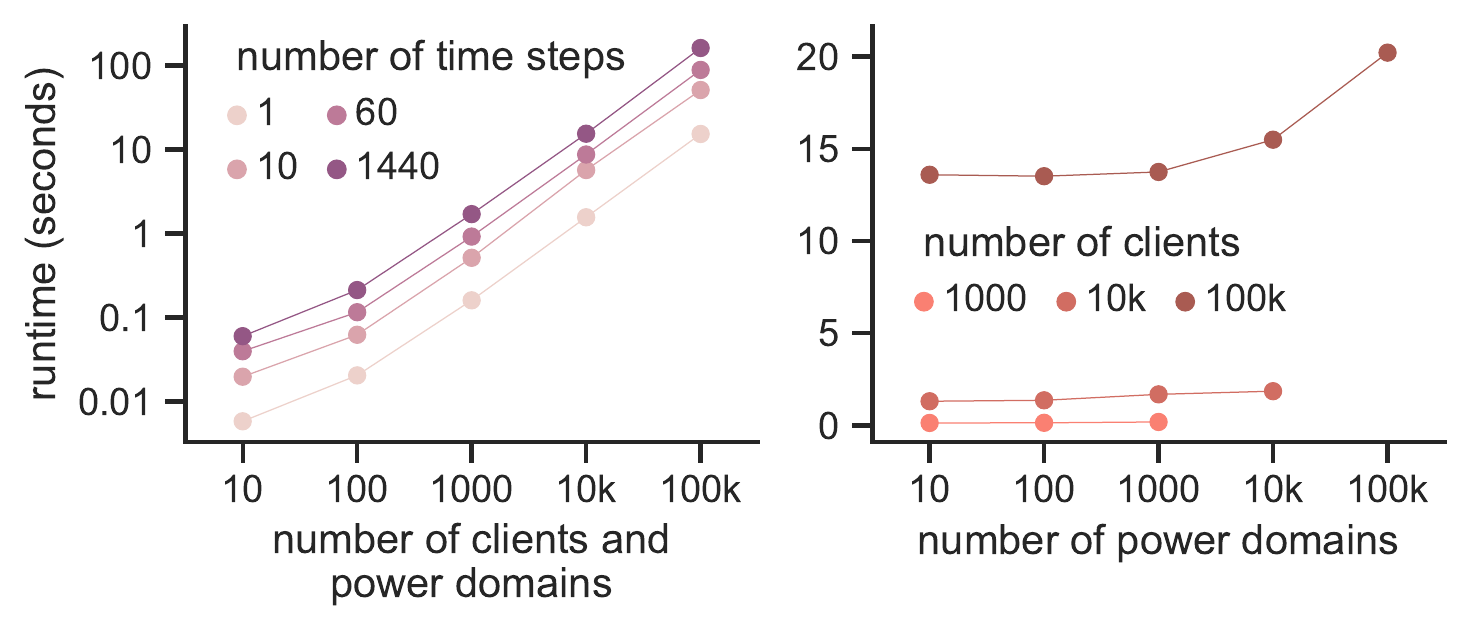}
       
   	\vspace{-6mm}\hspace{.1\linewidth}\subfloat[\label{fig:scalability-a} Influence of timestep search space.]{\hspace{.4\linewidth}}
   	\hspace{.05\linewidth}
	\subfloat[\label{fig:scalability-b} Influence of number of clients and power domains.]{\hspace{.4\linewidth}}
    \vspace{-2mm}
    \caption{FedZero overhead analysis.}
    \label{fig:scalability}
\end{figure}

We analyze the overhead of FedZero's client selection by profiling the runtime of Algorithm~\ref{alg:1}, including the MIP, on an Apple M1 processor.
Each experiment was repeated 5 times and we report mean values.
Figure~\ref{fig:scalability-a} shows the linear growth of runtime in regards to the number of clients: Even at the biggest evaluated setting, 100k clients distributed over 100k power domains searching over 1440 timesteps (24 hours in 1-minute resolution), the algorithm returns within two minutes.
For scenarios in the scale of the previous evaluation (100 clients, 10 power domains, 60 timesteps) it only takes around 0.1 seconds to decide on a set of clients.
We observe that due to the $\mathcal{O}(\log{}n)$ runtime of the binary search, increasing the timestep search space from 60 to 1440 (factor 24) only increases the runtime by factor 1.8.
Figure~\ref{fig:scalability-b} shows the runtime of a single MIP for different numbers of clients and power domains (note, that the y-axis is linear in this figure).
We observe, that the number of power domains has little to no impact on the runtime for up to 10k domains. Increasing the number of power domains from 10k to 100k only increases the runtime from 15.4 to 20.1 seconds.

In terms of communication overhead, FedZero requires excess energy and load forecasts at the beginning of each round.
Furthermore, clients periodically (in our evaluation minutely) sync with their power domain to align their performance with the actual available excess energy at training time.
As both payloads are in the order of kilobytes and we are assuming clients are connected via fiber, this overhead is negligible.

\section{Related Work}
\label{sec:rl}

\parag{Carbon-aware computing} 
As carbon pricing mechanisms, such as emission trading systems or carbon taxes, are starting to be implemented around the globe~\cite{WorldBank_CarbonPricing_2022}, the IT industry is pushing to increase the usage of low-carbon energy in datacenters.
Carbon-aware computing tries to reduce the emissions associated with computing by shifting flexible workloads towards times~\cite{Wiesner_LetsWaitAwhile_2021, Radovanovic_Google_2021, Fridgen_NotAllDoomAndGloom_2021} and locations~\cite{Zheng_MitigatingCurtailment_2020, Zhou_CarbonAwareLoadBalancing_2013} with clean energy.
For example, Google defers delay-tolerant workloads when power is associated with high carbon intensity as a measure to reach their 24/7 carbon-free target by 2030~\cite{Radovanovic_Google_2021}.

While most research in carbon-aware computing aims at consuming cleaner energy from the public grid~\cite{Radovanovic_Google_2021, Lin_CouplingDatacentersPowerGrids_2021, Wiesner_LetsWaitAwhile_2021, Zhou_CarbonAwareLoadBalancing_2013}, recent works also try to better exploit excess energy, similar to FedZero.
Cucumber~\cite{Wiesner_Cucumber_2022} is an admission control policy that accepts low-priority workloads on underutilized infrastructure, only if they can be computed using excess energy.
Similarly, Zheng et al.~\cite{Zheng_MitigatingCurtailment_2020} explore workload migration on underutilized data centers as a measure to reduce curtailment.
The Zero-Carbon Cloud~\cite{chien2019zero} already targets the problem of curtailment at the level of infrastructure planning, by placing data centers close to renewable energy sources.

\parag{Carbon footprint of ML} 
The training of large ML models is a highly relevant domain for carbon-awareness, due to the often excessive energy requirements on the one hand, and certain flexibility in scheduling on the other.
Other than inference, which is usually expected to happen at low latency, ML training jobs can often be stopped and resumed, scaled up or down, or even migrated between locations.
Because of this, many papers have previously addressed the carbon emissions of centralized ML~\cite{Dodge_CarbonIntensityAICloudInstances_2022, Google_CarbonFootprintMLShrink_2022, Strubell_EnergyConsiderationsDLResearch_2020, Payal_NatureCarbonImpactAI_2020}.

Qiu et al.~\cite{Qiu_FirstLookCarbonFootprintFL_2021} were the first to broadly study the energy consumption and carbon footprint of FL and state that, "depending on the configuration, FL can emit up to two orders of magnitude more carbon than centralized machine learning."
Further studies investigate the carbon impact of hyperparameters such as concurrency rate~\cite{greenfl_2023} or the cost of differential privacy~\cite{energy_privacy_fl_eenergy_2023}.
Carbon awareness in the context of FL has so far only been explored by Güler and Yener~\cite{Guler_SustainableFederatedLearning_2021} who define a model for intermittent energy arrivals and propose a scheduler with provable convergence guarantees.
However, their assumptions regarding energy arrivals are highly simplified as they neither consider spare capacity on clients nor non-iid data distributions.
Moreover, other than FedZero, their model does not allow multiple clients to share the same power domain.

\parag{FL client selection}
Active (or guided) client selection in FL has received significant attention in recent years, as researchers try to improve the final accuracy, convergence speed, reliability, fairness, or reduce communication overhead compared to random client selection~\cite{Soltani_SurveyParticipantSelection_2022}.
For example, Oort~\cite{Lai_Oort_2021} exploits heterogeneous device capabilities and data characteristics by cherry-picking clients with high statistical model efficiency as well as high system utility, resulting in faster convergence and better final accuracy.
Similarly, other novel approaches like FedMarl~\cite{Zhang_FedMarl_2022}
or Power-of-Choice~\cite{JeeCho_PowerOfChoice_2022} utilize the local training loss of clients to bias the selection.
FedZero does not try to compete with but can be adapted to integrate with other client selection strategies.

Few energy-aware client selection strategies exist, like EAFL~\cite{Arouj_EAFL_2022}, that extend Oort's utility function to additionally consider the battery level of clients.
However, like most research addressing energy usage in FL~\cite{Yang_EnergyEfficientFLWireless_2021, Wang_EnergyMinimizationFLBatteryPowered_2022}, it aims to increase the operating time of battery-constrained end devices and is not concerned with the overall emissions associated with the training.
FedZero describes the first approach for FL training solely on excess energy and spare computing capacity.

\section{Conclusion}
\label{sec:conclusion}

This paper proposes FedZero, a system design for fast, efficient, and fair training of FL models using only renewable excess energy and spare computing capacity.
Our results show, that FedZero's client selection strategy converges significantly faster than all baselines under the mentioned resource constraints while ensuring fair participation of clients, even under highly imbalanced conditions.
Moreover, our approach is robust against forecasting errors and scalable to large-scale, globally distributed scenarios.

In future work, we want to investigate the impact of periodic patterns in excess energy availability on training performance~\cite{diurnal_or_nocturnal_ICLR_2022}.
Furthermore, we plan to integrate FedZero with novel asynchronous~\cite{pisces_socc22} or semi-synchronous~\cite{REFL_Eurosys23} strategies, while explicitly taking energy storage and grid carbon intensity into account.

\begin{acks}
We sincerely thank Solcast for providing us with free access to their solar forecast APIs.
We furthermore want to thank the anonymous reviewers of ICDCS '23 and e-Energy '24 for their helpful comments.
This research was supported by the German Academic Exchange Service (DAAD) as ide3a and IFI as well as the German Ministry for~Education and Research (BMBF) as \mbox{BIFOLD} (grant 01IS18025A) and Software Campus (grant 01IS17050).
\end{acks}

\bibliographystyle{ACM-Reference-Format}
\bibliography{bibliography}


\begin{thebibliography}{68}


\ifx \showCODEN    \undefined \def \showCODEN     #1{\unskip}     \fi
\ifx \showDOI      \undefined \def \showDOI       #1{#1}\fi
\ifx \showISBNx    \undefined \def \showISBNx     #1{\unskip}     \fi
\ifx \showISBNxiii \undefined \def \showISBNxiii  #1{\unskip}     \fi
\ifx \showISSN     \undefined \def \showISSN      #1{\unskip}     \fi
\ifx \showLCCN     \undefined \def \showLCCN      #1{\unskip}     \fi
\ifx \shownote     \undefined \def \shownote      #1{#1}          \fi
\ifx \showarticletitle \undefined \def \showarticletitle #1{#1}   \fi
\ifx \showURL      \undefined \def \showURL       {\relax}        \fi
\providecommand\bibfield[2]{#2}
\providecommand\bibinfo[2]{#2}
\providecommand\natexlab[1]{#1}
\providecommand\showeprint[2][]{arXiv:#2}

\bibitem[Abdelmoniem et~al\mbox{.}(2023)]%
        {REFL_Eurosys23}
\bibfield{author}{\bibinfo{person}{Ahmed~M. Abdelmoniem}, \bibinfo{person}{Atal~Narayan Sahu}, \bibinfo{person}{Marco Canini}, {and} \bibinfo{person}{Suhaib~A. Fahmy}.} \bibinfo{year}{2023}\natexlab{}.
\newblock \showarticletitle{REFL: Resource-Efficient Federated Learning}. In \bibinfo{booktitle}{\emph{EuroSys}}. \bibinfo{publisher}{ACM}.
\newblock
\showISBNx{9781450394871}
\urldef\tempurl%
\url{https://doi.org/10.1145/3552326.3567485}
\showDOI{\tempurl}


\bibitem[Alencar et~al\mbox{.}(2017)]%
        {Alencar2017DifferentMFwindpowergenerationcasestudy}
\bibfield{author}{\bibinfo{person}{David~B. Alencar}, \bibinfo{person}{Carolina de Mattos~Affonso}, \bibinfo{person}{Roberto C.~L. Oliveira}, \bibinfo{person}{Jorge Laureano~Moya Rodr{\'i}guez}, \bibinfo{person}{Jandecy~Cabral Leite}, {and} \bibinfo{person}{Jose Carlos~R. Filho}.} \bibinfo{year}{2017}\natexlab{}.
\newblock \showarticletitle{Different Models for Forecasting Wind Power Generation: Case Study}.
\newblock \bibinfo{journal}{\emph{Energies}}  \bibinfo{volume}{10} (\bibinfo{year}{2017}).
\newblock
\urldef\tempurl%
\url{https://doi.org/10.3390/en10121976}
\showDOI{\tempurl}


\bibitem[Amazon(2022)]%
        {AmazonCarbonFree}
\bibfield{author}{\bibinfo{person}{Amazon}.} \bibinfo{year}{2022}\natexlab{}.
\newblock \showarticletitle{Amazon's 2022 Sustainability Report}.
\newblock  (\bibinfo{year}{2022}).
\newblock


\bibitem[Arouj and Abdelmoniem(2022)]%
        {Arouj_EAFL_2022}
\bibfield{author}{\bibinfo{person}{Amna Arouj} {and} \bibinfo{person}{Ahmed~M. Abdelmoniem}.} \bibinfo{year}{2022}\natexlab{}.
\newblock \showarticletitle{Towards Energy-Aware Federated Learning on Battery-Powered Clients}. In \bibinfo{booktitle}{\emph{Workshop on Data Privacy and Federated Learning Technologies for Mobile Edge Network at ACM MobiCom}}.
\newblock


\bibitem[Bank(2022)]%
        {WorldBank_CarbonPricing_2022}
\bibfield{author}{\bibinfo{person}{World Bank}.} \bibinfo{year}{2022}\natexlab{}.
\newblock \bibinfo{booktitle}{\emph{State and Trends of Carbon Pricing 2022}}.
\newblock \bibinfo{type}{{T}echnical {R}eport}. \bibinfo{institution}{Washington, DC: World Bank.}
\newblock


\bibitem[Bashir et~al\mbox{.}(2022)]%
        {Bashir_SustCompWithoutHotAir_2022}
\bibfield{author}{\bibinfo{person}{Noman Bashir}, \bibinfo{person}{David Irwin}, \bibinfo{person}{Prashant Shenoy}, {and} \bibinfo{person}{Abel Souza}.} \bibinfo{year}{2022}\natexlab{}.
\newblock \showarticletitle{Sustainable Computing - Without the Hot Air}. In \bibinfo{booktitle}{\emph{HotCarbon}}.
\newblock


\bibitem[Berg et~al\mbox{.}(2021)]%
        {berg21_interspeech}
\bibfield{author}{\bibinfo{person}{Axel Berg}, \bibinfo{person}{Mark O’Connor}, {and} \bibinfo{person}{Miguel~Tairum Cruz}.} \bibinfo{year}{2021}\natexlab{}.
\newblock \showarticletitle{{Keyword Transformer: A Self-Attention Model for Keyword Spotting}}. In \bibinfo{booktitle}{\emph{Proc. Interspeech 2021}}. \bibinfo{pages}{4249--4253}.
\newblock
\urldef\tempurl%
\url{https://doi.org/10.21437/Interspeech.2021-1286}
\showDOI{\tempurl}


\bibitem[Beutel et~al\mbox{.}(2020)]%
        {Beutel_Flower_2020}
\bibfield{author}{\bibinfo{person}{Daniel~J Beutel}, \bibinfo{person}{Taner Topal}, \bibinfo{person}{Akhil Mathur}, \bibinfo{person}{Xinchi Qiu}, \bibinfo{person}{Titouan Parcollet}, {and} \bibinfo{person}{Nicholas~D Lane}.} \bibinfo{year}{2020}\natexlab{}.
\newblock \showarticletitle{Flower: A Friendly Federated Learning Research Framework}.
\newblock \bibinfo{journal}{\emph{arXiv preprint arXiv:2007.14390}} (\bibinfo{year}{2020}).
\newblock


\bibitem[Bj{\o}rn et~al\mbox{.}(2022)]%
        {Bjorn_RECThreatenTargetsNature_2022}
\bibfield{author}{\bibinfo{person}{Anders Bj{\o}rn}, \bibinfo{person}{Shannon~M. Lloyd}, \bibinfo{person}{Matthew Brander}, {and} \bibinfo{person}{H.~Damon Matthews}.} \bibinfo{year}{2022}\natexlab{}.
\newblock \showarticletitle{Renewable energy certificates threaten the integrity of corporate science-based targets}.
\newblock \bibinfo{journal}{\emph{Nature Climate Change}} \bibinfo{volume}{12}, \bibinfo{number}{6} (\bibinfo{year}{2022}).
\newblock


\bibitem[Bonawitz et~al\mbox{.}(2019)]%
        {Bonawitz_FlAtScale_2019}
\bibfield{author}{\bibinfo{person}{Keith Bonawitz}, \bibinfo{person}{Hubert Eichner}, \bibinfo{person}{Wolfgang Grieskamp}, \bibinfo{person}{Dzmitry Huba}, \bibinfo{person}{Alex Ingerman}, \bibinfo{person}{Vladimir Ivanov}, \bibinfo{person}{Chlo\'{e} Kiddon}, \bibinfo{person}{Jakub Kone\v{c}n\'{y}}, \bibinfo{person}{Stefano Mazzocchi}, \bibinfo{person}{Brendan McMahan}, \bibinfo{person}{Timon Van~Overveldt}, \bibinfo{person}{David Petrou}, \bibinfo{person}{Daniel Ramage}, {and} \bibinfo{person}{Jason Roselander}.} \bibinfo{year}{2019}\natexlab{}.
\newblock \showarticletitle{Towards Federated Learning at Scale: System Design}. In \bibinfo{booktitle}{\emph{MLSys}}.
\newblock
\urldef\tempurl%
\url{https://proceedings.mlsys.org/paper/2019/file/bd686fd640be98efaae0091fa301e613-Paper.pdf}
\showURL{%
\tempurl}


\bibitem[Bright et~al\mbox{.}(2018)]%
        {BRIGHT2018118satellitederived}
\bibfield{author}{\bibinfo{person}{Jamie~M. Bright}, \bibinfo{person}{Sven Killinger}, \bibinfo{person}{David Lingfors}, {and} \bibinfo{person}{Nicholas~A. Engerer}.} \bibinfo{year}{2018}\natexlab{}.
\newblock \showarticletitle{Improved satellite-derived PV power nowcasting using real-time power data from reference PV systems}.
\newblock \bibinfo{journal}{\emph{Solar Energy}}  \bibinfo{volume}{168} (\bibinfo{year}{2018}).
\newblock
\urldef\tempurl%
\url{https://doi.org/10.1016/j.solener.2017.10.091}
\showDOI{\tempurl}


\bibitem[Caldas et~al\mbox{.}(2019)]%
        {Caldas_LEAF_2019}
\bibfield{author}{\bibinfo{person}{Sebastian Caldas}, \bibinfo{person}{Sai Meher Karthik~Duddu}, \bibinfo{person}{Peter Wu}, \bibinfo{person}{Tian Li}, \bibinfo{person}{Jakub Kone{\v{c}}n{\`y}}, \bibinfo{person}{H~Brendan McMahan}, \bibinfo{person}{Virginia Smith}, {and} \bibinfo{person}{Ameet Talwalkar}.} \bibinfo{year}{2019}\natexlab{}.
\newblock \showarticletitle{{LEAF}: A Benchmark for Federated Settings}. In \bibinfo{booktitle}{\emph{Workshop on Federated Learning for Data Privacy and Confidentiality at NeurIPS}}.
\newblock


\bibitem[{California ISO}(2024)]%
        {CAISO_ManagingOversupply_2022}
\bibfield{author}{\bibinfo{person}{{California ISO}}.} \bibinfo{year}{2024}\natexlab{}.
\newblock \bibinfo{title}{Managing oversupply}.
\newblock \bibinfo{howpublished}{\url{http://www.caiso.com/informed/Pages/ManagingOversupply.aspx}}.
\newblock
\newblock
\shownote{accessed Jan. 2024}.


\bibitem[Chien et~al\mbox{.}(2022)]%
        {Chien_BeyondPUE_2022}
\bibfield{author}{\bibinfo{person}{Andrew Chien}, \bibinfo{person}{Chaojie Zhang}, \bibinfo{person}{Liuzixuan Lin}, {and} \bibinfo{person}{Varsha Rao}.} \bibinfo{year}{2022}\natexlab{}.
\newblock \showarticletitle{Beyond {PUE}: Flexible Datacenters Empowering the Cloud to Decarbonize}. In \bibinfo{booktitle}{\emph{HotCarbon}}.
\newblock


\bibitem[Chien et~al\mbox{.}(2019)]%
        {chien2019zero}
\bibfield{author}{\bibinfo{person}{Andrew~A Chien}, \bibinfo{person}{Chaojie Zhang}, {and} \bibinfo{person}{Hai~Duc Nguyen}.} \bibinfo{year}{2019}\natexlab{}.
\newblock \showarticletitle{Zero-carbon Cloud: Research Challenges for Datacenters as Supply-following Loads}.
\newblock \bibinfo{journal}{\emph{University of Chicago, Tech. Rep. CS-TR-2019-08}} (\bibinfo{year}{2019}).
\newblock


\bibitem[Dhar(2020)]%
        {Payal_NatureCarbonImpactAI_2020}
\bibfield{author}{\bibinfo{person}{Payal Dhar}.} \bibinfo{year}{2020}\natexlab{}.
\newblock \showarticletitle{The carbon impact of artificial intelligence}.
\newblock \bibinfo{journal}{\emph{Nature Machine Intelligence}}  \bibinfo{volume}{2} (\bibinfo{year}{2020}), \bibinfo{pages}{423--425}.
\newblock


\bibitem[Dodge et~al\mbox{.}(2022)]%
        {Dodge_CarbonIntensityAICloudInstances_2022}
\bibfield{author}{\bibinfo{person}{Jesse Dodge}, \bibinfo{person}{Taylor Prewitt}, \bibinfo{person}{Remi Tachet~des Combes}, \bibinfo{person}{Erika Odmark}, \bibinfo{person}{Roy Schwartz}, \bibinfo{person}{Emma Strubell}, \bibinfo{person}{Alexandra~Sasha Luccioni}, \bibinfo{person}{Noah~A. Smith}, \bibinfo{person}{Nicole DeCario}, {and} \bibinfo{person}{Will Buchanan}.} \bibinfo{year}{2022}\natexlab{}.
\newblock \showarticletitle{Measuring the Carbon Intensity of {AI} in Cloud Instances}. In \bibinfo{booktitle}{\emph{ACM FAccT}}.
\newblock
\showISBNx{9781450393522}
\urldef\tempurl%
\url{https://doi.org/10.1145/3531146.3533234}
\showDOI{\tempurl}


\bibitem[Enes et~al\mbox{.}(2020)]%
        {Enes_PowerBudgetingCluster_2020}
\bibfield{author}{\bibinfo{person}{Jonatan Enes}, \bibinfo{person}{Guillaume Fieni}, \bibinfo{person}{Roberto~R. Expósito}, \bibinfo{person}{Romain Rouvoy}, {and} \bibinfo{person}{Juan Touriño}.} \bibinfo{year}{2020}\natexlab{}.
\newblock \showarticletitle{Power Budgeting of Big Data Applications in Container-based Clusters}. In \bibinfo{booktitle}{\emph{IEEE CLUSTER}}.
\newblock


\bibitem[Fridgen et~al\mbox{.}(2021)]%
        {Fridgen_NotAllDoomAndGloom_2021}
\bibfield{author}{\bibinfo{person}{Gilbert Fridgen}, \bibinfo{person}{Marc-Fabian K{\"o}rner}, \bibinfo{person}{Steffen Walters}, {and} \bibinfo{person}{Martin Weibelzahl}.} \bibinfo{year}{2021}\natexlab{}.
\newblock \showarticletitle{Not All Doom and Gloom: How Energy-Intensive and Temporally Flexible Data Center Applications May Actually Promote Renewable Energy Sources}.
\newblock \bibinfo{journal}{\emph{Business {\&} Information Systems Engineering}} \bibinfo{volume}{63}, \bibinfo{number}{3} (\bibinfo{year}{2021}).
\newblock


\bibitem[Google(2022)]%
        {GoogleCarbonFree}
\bibfield{author}{\bibinfo{person}{Google}.} \bibinfo{year}{2022}\natexlab{}.
\newblock \showarticletitle{2022 {E}nvironmental Report}.
\newblock  (\bibinfo{year}{2022}).
\newblock


\bibitem[Güler and Yener(2021)]%
        {Guler_SustainableFederatedLearning_2021}
\bibfield{author}{\bibinfo{person}{Başak Güler} {and} \bibinfo{person}{Aylin Yener}.} \bibinfo{year}{2021}\natexlab{}.
\newblock \showarticletitle{A Framework for Sustainable Federated Learning}. In \bibinfo{booktitle}{\emph{2021 19th International Symposium on Modeling and Optimization in Mobile, Ad hoc, and Wireless Networks (WiOpt)}}.
\newblock
\urldef\tempurl%
\url{https://doi.org/10.23919/WiOpt52861.2021.9589930}
\showDOI{\tempurl}


\bibitem[Hsu et~al\mbox{.}(2019)]%
        {noniid_dirichlet}
\bibfield{author}{\bibinfo{person}{Harry Hsu}, \bibinfo{person}{Hang Qi}, {and} \bibinfo{person}{Matthew Brown}.} \bibinfo{year}{2019}\natexlab{}.
\newblock \showarticletitle{Measuring the Effects of Non-Identical Data Distribution for Federated Visual Classification}.
\newblock \bibinfo{journal}{\emph{arXiv preprint arXiv:1909.06335}} (\bibinfo{year}{2019}).
\newblock


\bibitem[Huang et~al\mbox{.}(2017)]%
        {DenseNet_2017}
\bibfield{author}{\bibinfo{person}{Gao Huang}, \bibinfo{person}{Zhuang Liu}, \bibinfo{person}{Laurens Van Der~Maaten}, {and} \bibinfo{person}{Kilian~Q. Weinberger}.} \bibinfo{year}{2017}\natexlab{}.
\newblock \showarticletitle{Densely Connected Convolutional Networks}. In \bibinfo{booktitle}{\emph{CVPR}}.
\newblock


\bibitem[Jee~Cho et~al\mbox{.}(2022)]%
        {JeeCho_PowerOfChoice_2022}
\bibfield{author}{\bibinfo{person}{Yae Jee~Cho}, \bibinfo{person}{Jianyu Wang}, {and} \bibinfo{person}{Gauri Joshi}.} \bibinfo{year}{2022}\natexlab{}.
\newblock \showarticletitle{Towards Understanding Biased Client Selection in Federated Learning}. In \bibinfo{booktitle}{\emph{AISTATS}}.
\newblock


\bibitem[Jiang et~al\mbox{.}(2020)]%
        {Jiang_FLSmartCity_2020}
\bibfield{author}{\bibinfo{person}{Ji~Chu Jiang}, \bibinfo{person}{Burak Kantarci}, \bibinfo{person}{Sema Oktug}, {and} \bibinfo{person}{Tolga Soyata}.} \bibinfo{year}{2020}\natexlab{}.
\newblock \showarticletitle{Federated Learning in Smart City Sensing: Challenges and Opportunities}.
\newblock \bibinfo{journal}{\emph{Sensors}} \bibinfo{volume}{20}, \bibinfo{number}{21} (\bibinfo{year}{2020}).
\newblock


\bibitem[Jiang et~al\mbox{.}(2022)]%
        {pisces_socc22}
\bibfield{author}{\bibinfo{person}{Zhifeng Jiang}, \bibinfo{person}{Wei Wang}, \bibinfo{person}{Baochun Li}, {and} \bibinfo{person}{Bo Li}.} \bibinfo{year}{2022}\natexlab{}.
\newblock \showarticletitle{Pisces: Efficient Federated Learning via Guided Asynchronous Training}. In \bibinfo{booktitle}{\emph{ACM Symposium on Cloud Computing (SoCC)}}.
\newblock
\urldef\tempurl%
\url{https://doi.org/10.1145/3542929.3563463}
\showDOI{\tempurl}


\bibitem[Joppa(2021)]%
        {Microsoft_CarbonCounting_2021}
\bibfield{author}{\bibinfo{person}{Lucas Joppa}.} \bibinfo{year}{2021}\natexlab{}.
\newblock \bibinfo{title}{Made to measure: Sustainability commitment progress and updates}.
\newblock \bibinfo{howpublished}{Microsoft}.
\newblock
\urldef\tempurl%
\url{https://blogs.microsoft.com/blog/2021/07/14/made-to-measure-sustainability-commitment-progress-and-updates}
\showURL{%
Retrieved Sept. 2023 from \tempurl}


\bibitem[Khalyasmaa et~al\mbox{.}(2019)]%
        {Khalyasmaa2019PredictionOS}
\bibfield{author}{\bibinfo{person}{Alexandra~I. Khalyasmaa}, \bibinfo{person}{Stanislav~A. Eroshenko}, \bibinfo{person}{T. Chakravarthy}, \bibinfo{person}{Venu~Gopal Gasi}, \bibinfo{person}{Sandeep Kumar~Yadav Bollu}, \bibinfo{person}{Raphael Caire}, \bibinfo{person}{Sai Kumar~Reddy Atluri}, {and} \bibinfo{person}{Suresh Karrolla}.} \bibinfo{year}{2019}\natexlab{}.
\newblock \showarticletitle{Prediction of Solar Power Generation Based on Random Forest Regressor Model}. In \bibinfo{booktitle}{\emph{IEEE SIBIRCON}}.
\newblock
\urldef\tempurl%
\url{https://doi.org/10.1109/SIBIRCON48586.2019.8958063}
\showDOI{\tempurl}


\bibitem[Krizhevsky(2009)]%
        {Krizhevsky09learningmultiple}
\bibfield{author}{\bibinfo{person}{Alex Krizhevsky}.} \bibinfo{year}{2009}\natexlab{}.
\newblock \bibinfo{booktitle}{\emph{Learning multiple layers of features from tiny images}}.
\newblock \bibinfo{type}{{T}echnical {R}eport}.
\newblock


\bibitem[Lai et~al\mbox{.}(2021)]%
        {Lai_Oort_2021}
\bibfield{author}{\bibinfo{person}{Fan Lai}, \bibinfo{person}{Xiangfeng Zhu}, \bibinfo{person}{Harsha~V. Madhyastha}, {and} \bibinfo{person}{Mosharaf Chowdhury}.} \bibinfo{year}{2021}\natexlab{}.
\newblock \showarticletitle{Oort: Efficient Federated Learning via Guided Participant Selection}. In \bibinfo{booktitle}{\emph{USENIX OSDI}}.
\newblock
\showISBNx{978-1-939133-22-9}
\urldef\tempurl%
\url{https://www.usenix.org/conference/osdi21/presentation/lai}
\showURL{%
\tempurl}


\bibitem[Li et~al\mbox{.}(2022)]%
        {PyramidFL_2022}
\bibfield{author}{\bibinfo{person}{Chenning Li}, \bibinfo{person}{Xiao Zeng}, \bibinfo{person}{Mi Zhang}, {and} \bibinfo{person}{Zhichao Cao}.} \bibinfo{year}{2022}\natexlab{}.
\newblock \showarticletitle{PyramidFL: A Fine-Grained Client Selection Framework for Efficient Federated Learning}. In \bibinfo{booktitle}{\emph{ACM MobiCom}}.
\newblock
\urldef\tempurl%
\url{https://doi.org/10.1145/3495243.3517017}
\showDOI{\tempurl}


\bibitem[Li et~al\mbox{.}(2021)]%
        {LI2021121075powerpredictionwindturbinesgaussian}
\bibfield{author}{\bibinfo{person}{Qing'an Li}, \bibinfo{person}{Chang Cai}, \bibinfo{person}{Yasunari Kamada}, \bibinfo{person}{Takao Maeda}, \bibinfo{person}{Yuto Hiromori}, \bibinfo{person}{Shuni Zhou}, {and} \bibinfo{person}{Jianzhong Xu}.} \bibinfo{year}{2021}\natexlab{}.
\newblock \showarticletitle{Prediction of power generation of two 30 kW Horizontal Axis Wind Turbines with Gaussian model}.
\newblock \bibinfo{journal}{\emph{Energy}}  \bibinfo{volume}{231} (\bibinfo{year}{2021}).
\newblock
\urldef\tempurl%
\url{https://doi.org/10.1016/j.energy.2021.121075}
\showDOI{\tempurl}


\bibitem[Li et~al\mbox{.}(2020b)]%
        {Li_Thunderbolt_2020}
\bibfield{author}{\bibinfo{person}{Shaohong Li}, \bibinfo{person}{Xi Wang}, \bibinfo{person}{Xiao Zhang}, \bibinfo{person}{Vasileios Kontorinis}, \bibinfo{person}{Sreekumar Kodakara}, \bibinfo{person}{David Lo}, {and} \bibinfo{person}{Parthasarathy Ranganathan}.} \bibinfo{year}{2020}\natexlab{b}.
\newblock \showarticletitle{Thunderbolt: {Throughput-Optimized}, {Quality-of-Service-Aware} Power Capping at Scale}. In \bibinfo{booktitle}{\emph{USENIX OSDI}}.
\newblock
\showISBNx{ISBN 978-1-939133-19-9}
\urldef\tempurl%
\url{https://www.usenix.org/conference/osdi20/presentation/li-shaohong}
\showURL{%
\tempurl}


\bibitem[Li et~al\mbox{.}(2020a)]%
        {Li_FedProx_2020}
\bibfield{author}{\bibinfo{person}{Tian Li}, \bibinfo{person}{Anit~Kumar Sahu}, \bibinfo{person}{Manzil Zaheer}, \bibinfo{person}{Maziar Sanjabi}, \bibinfo{person}{Ameet Talwalkar}, {and} \bibinfo{person}{Virginia Smith}.} \bibinfo{year}{2020}\natexlab{a}.
\newblock \showarticletitle{Federated Optimization in Heterogeneous Networks}. In \bibinfo{booktitle}{\emph{MLSys}}.
\newblock


\bibitem[Lin et~al\mbox{.}(2021)]%
        {Lin_CouplingDatacentersPowerGrids_2021}
\bibfield{author}{\bibinfo{person}{Liuzixuan Lin}, \bibinfo{person}{Victor~M. Zavala}, {and} \bibinfo{person}{Andrew Chien}.} \bibinfo{year}{2021}\natexlab{}.
\newblock \showarticletitle{Evaluating Coupling Models for Cloud Datacenters and Power Grids}. In \bibinfo{booktitle}{\emph{ACM e-Energy}}.
\newblock
\showISBNx{9781450383332}
\urldef\tempurl%
\url{https://doi.org/10.1145/3447555.3464868}
\showDOI{\tempurl}


\bibitem[Liu et~al\mbox{.}(2017)]%
        {Liu_BatteryAging_2017}
\bibfield{author}{\bibinfo{person}{Longjun Liu}, \bibinfo{person}{Hongbin Sun}, \bibinfo{person}{Chao Li}, \bibinfo{person}{Tao Li}, \bibinfo{person}{Jingmin Xin}, {and} \bibinfo{person}{Nanning Zheng}.} \bibinfo{year}{2017}\natexlab{}.
\newblock \showarticletitle{Managing Battery Aging for High Energy Availability in Green Datacenters}.
\newblock \bibinfo{journal}{\emph{IEEE Transactions on Parallel and Distributed Systems}} \bibinfo{volume}{28}, \bibinfo{number}{12} (\bibinfo{year}{2017}).
\newblock
\urldef\tempurl%
\url{https://doi.org/10.1109/TPDS.2017.2712778}
\showDOI{\tempurl}


\bibitem[McMahan et~al\mbox{.}(2016)]%
        {McMahan_FL_2016}
\bibfield{author}{\bibinfo{person}{H.~B. McMahan}, \bibinfo{person}{Eider Moore}, \bibinfo{person}{Daniel Ramage}, \bibinfo{person}{Seth Hampson}, {and} \bibinfo{person}{Blaise~Ag{\"u}era y Arcas}.} \bibinfo{year}{2016}\natexlab{}.
\newblock \showarticletitle{Communication-Efficient Learning of Deep Networks from Decentralized Data}. In \bibinfo{booktitle}{\emph{AISTATS}}.
\newblock


\bibitem[Microsoft(2022)]%
        {MicrosoftCarbonFree}
\bibfield{author}{\bibinfo{person}{Microsoft}.} \bibinfo{year}{2022}\natexlab{}.
\newblock \showarticletitle{2022 {E}nvironmental Sustainability Report}.
\newblock  (\bibinfo{year}{2022}).
\newblock


\bibitem[Naidu et~al\mbox{.}(2021)]%
        {Naidu_CarbonEmissionsDifferentiallyPrivate_2021}
\bibfield{author}{\bibinfo{person}{Rakshit Naidu}, \bibinfo{person}{Harshita Diddee}, \bibinfo{person}{Ajinkya~K Mulay}, \bibinfo{person}{Aleti Vardhan}, \bibinfo{person}{Krithika Ramesh}, {and} \bibinfo{person}{Ahmed Zamzam}.} \bibinfo{year}{2021}\natexlab{}.
\newblock \showarticletitle{Towards Quantifying the Carbon Emissions of Differentially Private Machine Learning}. In \bibinfo{booktitle}{\emph{Workshop on Socially Responsible Machine Learning at ICML}}.
\newblock


\bibitem[Nguyen et~al\mbox{.}(2022)]%
        {Nguyen_FLAutonomousDriving_2022}
\bibfield{author}{\bibinfo{person}{Anh Nguyen}, \bibinfo{person}{Tuong Do}, \bibinfo{person}{Minh Tran}, \bibinfo{person}{Binh~X. Nguyen}, \bibinfo{person}{Chien Duong}, \bibinfo{person}{Tu Phan}, \bibinfo{person}{Erman Tjiputra}, {and} \bibinfo{person}{Quang~D. Tran}.} \bibinfo{year}{2022}\natexlab{}.
\newblock \showarticletitle{Deep Federated Learning for Autonomous Driving}. In \bibinfo{booktitle}{\emph{2022 IEEE Intelligent Vehicles Symposium (IV)}}.
\newblock


\bibitem[Oster(2022)]%
        {Amazon_CarbonCounting_2022}
\bibfield{author}{\bibinfo{person}{Jake Oster}.} \bibinfo{year}{2022}\natexlab{}.
\newblock \bibinfo{title}{How we count carbon emissions from electricity matters}.
\newblock \bibinfo{howpublished}{Amazon}.
\newblock
\urldef\tempurl%
\url{https://www.amazon.science/blog/how-we-count-carbon-emissions-from-electricity-matters}
\showURL{%
Retrieved Sept. 2023 from \tempurl}


\bibitem[Patterson et~al\mbox{.}(2022)]%
        {Google_CarbonFootprintMLShrink_2022}
\bibfield{author}{\bibinfo{person}{David Patterson}, \bibinfo{person}{Joseph Gonzalez}, \bibinfo{person}{Urs Hölzle}, \bibinfo{person}{Quoc Le}, \bibinfo{person}{Chen Liang}, \bibinfo{person}{Lluis-Miquel Munguia}, \bibinfo{person}{Daniel Rothchild}, \bibinfo{person}{David~R. So}, \bibinfo{person}{Maud Texier}, {and} \bibinfo{person}{Jeff Dean}.} \bibinfo{year}{2022}\natexlab{}.
\newblock \showarticletitle{The Carbon Footprint of Machine Learning Training Will Plateau, Then Shrink}.
\newblock \bibinfo{journal}{\emph{Computer}} \bibinfo{volume}{55}, \bibinfo{number}{7} (\bibinfo{year}{2022}).
\newblock
\urldef\tempurl%
\url{https://doi.org/10.1109/MC.2022.3148714}
\showDOI{\tempurl}


\bibitem[Qiu et~al\mbox{.}(2021)]%
        {Qiu_FirstLookCarbonFootprintFL_2021}
\bibfield{author}{\bibinfo{person}{Xinchi Qiu}, \bibinfo{person}{Titouan Parcollet}, \bibinfo{person}{Javier Fernandez-Marques}, \bibinfo{person}{Pedro Porto~Buarque de Gusmao}, \bibinfo{person}{Daniel~J. Beutel}, \bibinfo{person}{Taner Topal}, \bibinfo{person}{Akhil Mathur}, {and} \bibinfo{person}{Nicholas~D. Lane}.} \bibinfo{year}{2021}\natexlab{}.
\newblock \showarticletitle{A first look into the carbon footprint of federated learning}.
\newblock \bibinfo{journal}{\emph{arXiv preprint arXiv:2102.07627}} (\bibinfo{year}{2021}).
\newblock


\bibitem[Radovanovic et~al\mbox{.}(2022)]%
        {Radovanovic_Google_2021}
\bibfield{author}{\bibinfo{person}{Ana Radovanovic}, \bibinfo{person}{Ross Koningstein}, \bibinfo{person}{Ian Schneider}, \bibinfo{person}{Bokan Chen}, \bibinfo{person}{Alexandre Duarte}, \bibinfo{person}{Binz Roy}, \bibinfo{person}{Diyue Xiao}, \bibinfo{person}{Maya Haridasan}, \bibinfo{person}{Patrick Hung}, \bibinfo{person}{Nick Care}, \bibinfo{person}{Saurav Talukdar}, \bibinfo{person}{Eric Mullen}, \bibinfo{person}{Kendal Smith}, \bibinfo{person}{Mariellen Cottman}, {and} \bibinfo{person}{Walfredo Cirne}.} \bibinfo{year}{2022}\natexlab{}.
\newblock \showarticletitle{Carbon-Aware Computing for Datacenters}.
\newblock \bibinfo{journal}{\emph{IEEE Transactions on Power Systems}} (\bibinfo{year}{2022}).
\newblock


\bibitem[Rapp et~al\mbox{.}(2022)]%
        {Rapp_DISTREAL_2022}
\bibfield{author}{\bibinfo{person}{Martin Rapp}, \bibinfo{person}{Ramin Khalili}, \bibinfo{person}{Kilian Pfeiffer}, {and} \bibinfo{person}{Jörg Henkel}.} \bibinfo{year}{2022}\natexlab{}.
\newblock \showarticletitle{DISTREAL: Distributed Resource-Aware Learning in Heterogeneous Systems}. In \bibinfo{booktitle}{\emph{AAAI}}.
\newblock


\bibitem[REN21(2022)]%
        {REN21_RenewablesReport_2022}
\bibfield{author}{\bibinfo{person}{REN21}.} \bibinfo{year}{2022}\natexlab{}.
\newblock \showarticletitle{Renewables 2022 Global Status Report}.
\newblock  (\bibinfo{year}{2022}).
\newblock


\bibitem[Rieke et~al\mbox{.}(2020)]%
        {Rieke_FLHealth_2020}
\bibfield{author}{\bibinfo{person}{Nicola Rieke}, \bibinfo{person}{Jonny Hancox}, \bibinfo{person}{Wenqi Li}, \bibinfo{person}{Fausto Milletar{\`i}}, \bibinfo{person}{Holger~R. Roth}, \bibinfo{person}{Shadi Albarqouni}, \bibinfo{person}{Spyridon Bakas}, \bibinfo{person}{Mathieu~N. Galtier}, \bibinfo{person}{Bennett~A. Landman}, \bibinfo{person}{Klaus Maier-Hein}, \bibinfo{person}{S{\'e}bastien Ourselin}, \bibinfo{person}{Micah Sheller}, \bibinfo{person}{Ronald~M. Summers}, \bibinfo{person}{Andrew Trask}, \bibinfo{person}{Daguang Xu}, \bibinfo{person}{Maximilian Baust}, {and} \bibinfo{person}{M.~Jorge Cardoso}.} \bibinfo{year}{2020}\natexlab{}.
\newblock \showarticletitle{The future of digital health with federated learning}.
\newblock \bibinfo{journal}{\emph{npj Digital Medicine}} \bibinfo{volume}{3}, \bibinfo{number}{1} (\bibinfo{year}{2020}).
\newblock


\bibitem[Schwermer et~al\mbox{.}(2023)]%
        {energy_privacy_fl_eenergy_2023}
\bibfield{author}{\bibinfo{person}{Ren\'{e} Schwermer}, \bibinfo{person}{Ruben Mayer}, {and} \bibinfo{person}{Hans-Arno Jacobsen}.} \bibinfo{year}{2023}\natexlab{}.
\newblock \showarticletitle{Energy vs Privacy: Estimating the Ecological Impact of Federated Learning}. In \bibinfo{booktitle}{\emph{ACM e-Energy}}.
\newblock


\bibitem[So et~al\mbox{.}(2022)]%
        {So_FedSpace_2022}
\bibfield{author}{\bibinfo{person}{Jinhyun So}, \bibinfo{person}{Kevin Hsieh}, \bibinfo{person}{Behnaz Arzani}, \bibinfo{person}{Shadi Noghabi}, \bibinfo{person}{Salman Avestimehr}, {and} \bibinfo{person}{Ranveer Chandra}.} \bibinfo{year}{2022}\natexlab{}.
\newblock \showarticletitle{FedSpace: An Efficient Federated Learning Framework at Satellites and Ground Stations}.
\newblock \bibinfo{journal}{\emph{arXiv preprint arXiv:2202.01267}} (\bibinfo{year}{2022}).
\newblock


\bibitem[Soltani et~al\mbox{.}(2022)]%
        {Soltani_SurveyParticipantSelection_2022}
\bibfield{author}{\bibinfo{person}{Behnaz Soltani}, \bibinfo{person}{Venus Haghighi}, \bibinfo{person}{Adnan Mahmood}, \bibinfo{person}{Quan~Z. Sheng}, {and} \bibinfo{person}{Lina Yao}.} \bibinfo{year}{2022}\natexlab{}.
\newblock \showarticletitle{A Survey on Participant Selection for Federated Learning in Mobile Networks}. In \bibinfo{booktitle}{\emph{Workshop on Mobility in the Evolving Internet Architecture (MobiArch) at MobiCom}}.
\newblock


\bibitem[Souza et~al\mbox{.}(2023)]%
        {Souza_Ecovisor_2023}
\bibfield{author}{\bibinfo{person}{Abel Souza}, \bibinfo{person}{Noman Bashir}, \bibinfo{person}{Jorge Murillo}, \bibinfo{person}{Walid Hanafy}, \bibinfo{person}{Qianlin Liang}, \bibinfo{person}{David Irwin}, {and} \bibinfo{person}{Prashant Shenoy}.} \bibinfo{year}{2023}\natexlab{}.
\newblock \showarticletitle{Ecovisor: A Virtual Energy System for Carbon-Efficient Applications}. In \bibinfo{booktitle}{\emph{ASPLOS}}.
\newblock


\bibitem[Strubell et~al\mbox{.}(2020)]%
        {Strubell_EnergyConsiderationsDLResearch_2020}
\bibfield{author}{\bibinfo{person}{Emma Strubell}, \bibinfo{person}{Ananya Ganesh}, {and} \bibinfo{person}{Andrew McCallum}.} \bibinfo{year}{2020}\natexlab{}.
\newblock \showarticletitle{Energy and Policy Considerations for Modern Deep Learning Research}. In \bibinfo{booktitle}{\emph{AAAI}}.
\newblock


\bibitem[Tan and Le(2019)]%
        {pmlr-v97-tan19a}
\bibfield{author}{\bibinfo{person}{Mingxing Tan} {and} \bibinfo{person}{Quoc Le}.} \bibinfo{year}{2019}\natexlab{}.
\newblock \showarticletitle{{E}fficient{N}et: Rethinking Model Scaling for Convolutional Neural Networks}. In \bibinfo{booktitle}{\emph{ICML}}.
\newblock


\bibitem[Texier(2021)]%
        {Google_CarbonCounting_2021}
\bibfield{author}{\bibinfo{person}{Maud Texier}.} \bibinfo{year}{2021}\natexlab{}.
\newblock \bibinfo{title}{A timely new approach to certifying clean energy}.
\newblock \bibinfo{howpublished}{Google}.
\newblock
\urldef\tempurl%
\url{https://cloud.google.com/blog/topics/sustainability/t-eacs-offer-new-approach-to-certifying-clean-energy}
\showURL{%
Retrieved Sept. 2023 from \tempurl}


\bibitem[Tirmazi et~al\mbox{.}(2020)]%
        {Tirmazi_BorgNextGen_2020}
\bibfield{author}{\bibinfo{person}{Muhammad Tirmazi}, \bibinfo{person}{Adam Barker}, \bibinfo{person}{Nan Deng}, \bibinfo{person}{Md~E. Haque}, \bibinfo{person}{Zhijing~Gene Qin}, \bibinfo{person}{Steven Hand}, \bibinfo{person}{Mor Harchol-Balter}, {and} \bibinfo{person}{John Wilkes}.} \bibinfo{year}{2020}\natexlab{}.
\newblock \showarticletitle{Borg: The next Generation}. In \bibinfo{booktitle}{\emph{EuroSys}}.
\newblock


\bibitem[Wang et~al\mbox{.}(2022)]%
        {Wang_EnergyMinimizationFLBatteryPowered_2022}
\bibfield{author}{\bibinfo{person}{Cong Wang}, \bibinfo{person}{Bin Hu}, {and} \bibinfo{person}{Hongyi Wu}.} \bibinfo{year}{2022}\natexlab{}.
\newblock \showarticletitle{Energy Minimization for Federated Asynchronous Learning on Battery-Powered Mobile Devices via Application Co-running}. In \bibinfo{booktitle}{\emph{ICDCS}}.
\newblock


\bibitem[Weng et~al\mbox{.}(2022)]%
        {Weng_AlibabaTraces_2022}
\bibfield{author}{\bibinfo{person}{Qizhen Weng}, \bibinfo{person}{Wencong Xiao}, \bibinfo{person}{Yinghao Yu}, \bibinfo{person}{Wei Wang}, \bibinfo{person}{Cheng Wang}, \bibinfo{person}{Jian He}, \bibinfo{person}{Yong Li}, \bibinfo{person}{Liping Zhang}, \bibinfo{person}{Wei Lin}, {and} \bibinfo{person}{Yu Ding}.} \bibinfo{year}{2022}\natexlab{}.
\newblock \showarticletitle{{MLaaS} in the Wild: Workload Analysis and Scheduling in Large-Scale Heterogeneous {GPU} Clusters}. In \bibinfo{booktitle}{\emph{USENIX NSDI}}.
\newblock


\bibitem[Wiesner et~al\mbox{.}(2023)]%
        {wiesner2023vessim}
\bibfield{author}{\bibinfo{person}{Philipp Wiesner}, \bibinfo{person}{Ilja Behnke}, {and} \bibinfo{person}{Odej Kao}.} \bibinfo{year}{2023}\natexlab{}.
\newblock \bibinfo{title}{A Testbed for Carbon-Aware Applications and Systems}.
\newblock
\newblock
\showeprint[arxiv]{2306.09774}~[cs.DC]


\bibitem[Wiesner et~al\mbox{.}(2021)]%
        {Wiesner_LetsWaitAwhile_2021}
\bibfield{author}{\bibinfo{person}{Philipp Wiesner}, \bibinfo{person}{Ilja Behnke}, \bibinfo{person}{Dominik Scheinert}, \bibinfo{person}{Kordian Gontarska}, {and} \bibinfo{person}{Lauritz Thamsen}.} \bibinfo{year}{2021}\natexlab{}.
\newblock \showarticletitle{Let's Wait Awhile: How Temporal Workload Shifting Can Reduce Carbon Emissions in the Cloud}. In \bibinfo{booktitle}{\emph{ACM Middleware}}.
\newblock


\bibitem[Wiesner et~al\mbox{.}(2022)]%
        {Wiesner_Cucumber_2022}
\bibfield{author}{\bibinfo{person}{Philipp Wiesner}, \bibinfo{person}{Dominik Scheinert}, \bibinfo{person}{Thorsten Wittkopp}, \bibinfo{person}{Lauritz Thamsen}, {and} \bibinfo{person}{Odej Kao}.} \bibinfo{year}{2022}\natexlab{}.
\newblock \showarticletitle{Cucumber: Renewable-Aware Admission Control for Delay-Tolerant Cloud and Edge Workloads}. In \bibinfo{booktitle}{\emph{International European Conference on Parallel and Distributed Computing (Euro-Par)}}.
\newblock


\bibitem[Wu et~al\mbox{.}(2022)]%
        {Wu_SustainableAIMLSys_2022}
\bibfield{author}{\bibinfo{person}{Carole{-}Jean Wu}, \bibinfo{person}{Ramya Raghavendra}, \bibinfo{person}{Udit Gupta}, \bibinfo{person}{Bilge Acun}, \bibinfo{person}{Newsha Ardalani}, \bibinfo{person}{Kiwan Maeng}, \bibinfo{person}{Gloria Chang}, \bibinfo{person}{Fiona~Aga Behram}, \bibinfo{person}{Jinshi Huang}, \bibinfo{person}{Charles Bai}, \bibinfo{person}{Michael Gschwind}, \bibinfo{person}{Anurag Gupta}, \bibinfo{person}{Myle Ott}, \bibinfo{person}{Anastasia Melnikov}, \bibinfo{person}{Salvatore Candido}, \bibinfo{person}{David Brooks}, \bibinfo{person}{Geeta Chauhan}, \bibinfo{person}{Benjamin Lee}, \bibinfo{person}{Hsien{-}Hsin~S. Lee}, \bibinfo{person}{Bugra Akyildiz}, \bibinfo{person}{Maximilian Balandat}, \bibinfo{person}{Joe Spisak}, \bibinfo{person}{Ravi Jain}, \bibinfo{person}{Mike Rabbat}, {and} \bibinfo{person}{Kim~M. Hazelwood}.} \bibinfo{year}{2022}\natexlab{}.
\newblock \showarticletitle{Sustainable {AI:} Environmental Implications, Challenges and Opportunities}. In \bibinfo{booktitle}{\emph{MLSys}}.
\newblock


\bibitem[Yang et~al\mbox{.}(2019)]%
        {Yang_FLMorganClaypool_2019}
\bibfield{author}{\bibinfo{person}{Qiang Yang}, \bibinfo{person}{Yang Liu}, \bibinfo{person}{Yong Cheng}, \bibinfo{person}{Yan Kang}, \bibinfo{person}{Tianjian Chen}, {and} \bibinfo{person}{Han Yu}.} \bibinfo{year}{2019}\natexlab{}.
\newblock \bibinfo{booktitle}{\emph{Federated learning}}.
\newblock \bibinfo{publisher}{Morgan \& Claypool Publishers}.
\newblock


\bibitem[Yang et~al\mbox{.}(2021)]%
        {Yang_EnergyEfficientFLWireless_2021}
\bibfield{author}{\bibinfo{person}{Zhaohui Yang}, \bibinfo{person}{Mingzhe Chen}, \bibinfo{person}{Walid Saad}, \bibinfo{person}{Choong~Seon Hong}, {and} \bibinfo{person}{Mohammad Shikh-Bahaei}.} \bibinfo{year}{2021}\natexlab{}.
\newblock \showarticletitle{Energy Efficient Federated Learning Over Wireless Communication Networks}.
\newblock \bibinfo{journal}{\emph{IEEE Transactions on Wireless Communications}} \bibinfo{volume}{20}, \bibinfo{number}{3} (\bibinfo{year}{2021}).
\newblock


\bibitem[Yousefpour et~al\mbox{.}(2023)]%
        {greenfl_2023}
\bibfield{author}{\bibinfo{person}{Ashkan Yousefpour}, \bibinfo{person}{Shen Guo}, \bibinfo{person}{Ashish Shenoy}, \bibinfo{person}{Sayan Ghosh}, \bibinfo{person}{Pierre Stock}, \bibinfo{person}{Kiwan Maeng}, \bibinfo{person}{Schalk-Willem Krüger}, \bibinfo{person}{Michael Rabbat}, \bibinfo{person}{Carole-Jean Wu}, {and} \bibinfo{person}{Ilya Mironov}.} \bibinfo{year}{2023}\natexlab{}.
\newblock \bibinfo{title}{Green Federated Learning}.
\newblock
\newblock
\showeprint[arxiv]{2303.14604}~[cs.LG]


\bibitem[Zhang et~al\mbox{.}(2022)]%
        {Zhang_FedMarl_2022}
\bibfield{author}{\bibinfo{person}{Sai~Qian Zhang}, \bibinfo{person}{Jieyu Lin}, {and} \bibinfo{person}{Qi Zhang}.} \bibinfo{year}{2022}\natexlab{}.
\newblock \showarticletitle{A Multi-Agent Reinforcement Learning Approach for Efficient Client Selection in Federated Learning}. In \bibinfo{booktitle}{\emph{AAAI}}.
\newblock
\urldef\tempurl%
\url{https://doi.org/10.1609/aaai.v36i8.20894}
\showDOI{\tempurl}


\bibitem[Zheng et~al\mbox{.}(2020)]%
        {Zheng_MitigatingCurtailment_2020}
\bibfield{author}{\bibinfo{person}{Jiajia Zheng}, \bibinfo{person}{Andrew~A. Chien}, {and} \bibinfo{person}{Sangwon Suh}.} \bibinfo{year}{2020}\natexlab{}.
\newblock \showarticletitle{Mitigating Curtailment and Carbon Emissions through Load Migration between Data Centers}.
\newblock \bibinfo{journal}{\emph{Joule}} \bibinfo{volume}{4}, \bibinfo{number}{10} (\bibinfo{year}{2020}).
\newblock


\bibitem[Zhou et~al\mbox{.}(2013)]%
        {Zhou_CarbonAwareLoadBalancing_2013}
\bibfield{author}{\bibinfo{person}{Zhi Zhou}, \bibinfo{person}{Fangming Liu}, \bibinfo{person}{Yong Xu}, \bibinfo{person}{Ruolan Zou}, \bibinfo{person}{Hong Xu}, \bibinfo{person}{John~C.S. Lui}, {and} \bibinfo{person}{Hai Jin}.} \bibinfo{year}{2013}\natexlab{}.
\newblock \showarticletitle{Carbon-Aware Load Balancing for Geo-distributed Cloud Services}. In \bibinfo{booktitle}{\emph{21st Int. Symposium on Modelling, Analysis and Simulation of Computer and Telecommunication Systems (MASCOTS)}}.
\newblock


\bibitem[Zhu et~al\mbox{.}(2022)]%
        {diurnal_or_nocturnal_ICLR_2022}
\bibfield{author}{\bibinfo{person}{Chen Zhu}, \bibinfo{person}{Zheng Xu}, \bibinfo{person}{Mingqing Chen}, \bibinfo{person}{Jakub Konečný}, \bibinfo{person}{Andrew Hard}, {and} \bibinfo{person}{Tom Goldstein}.} \bibinfo{year}{2022}\natexlab{}.
\newblock \showarticletitle{Diurnal or Nocturnal? Federated Learning of Multi-branch Networks from Periodically Shifting Distributions}. In \bibinfo{booktitle}{\emph{ICLR}}.
\newblock


\end{thebibliography}

\clearpage

\appendix
\section{Table with all results}
\label{sec:appendix}

Note, that the \emph{Upper Bound} baseline is not constrained by capacity or energy availability and therefore also uses grid energy.

\vspace{5mm}

    \begin{tabular}{ccccccccccc}
    \toprule
    \multirow{3}[4]{*}{Dataset \& model} & \multirow{3}[4]{*}{Approach} & \multicolumn{4}{c}{Global scenario} &    & \multicolumn{4}{c}{Co-located scenario} \\
\cmidrule{3-6}\cmidrule{8-11}       &    & Target & Best & Time-to- & Energy-to- &    & Target & Best & Time-to- & Energy-to- \\
       &    & accuracy & accuracy & accuracy & accuracy &    & accuracy & accuracy & accuracy & accuracy \\
    \midrule
       & Upper bound & \multirow{8}[2]{*}{64.7 \%} & 68.3 \% & 1.6 d & \cellcolor[rgb]{ .851,  .851,  .851}91.1 kWh &    & \multirow{8}[2]{*}{65.5 \%} & 68.3 \% & 2.0 d & \cellcolor[rgb]{ .851,  .851,  .851}117.5 kWh \\
       & Random &    & 64.7 \% & 6.7 d & \cellcolor[rgb]{ .886,  .937,  .855}80.6 kWh &    &    & 65.5 \% & 6.7 d & \cellcolor[rgb]{ .886,  .937,  .855}101.0 kWh \\
       & Random 1.3n &    & 66.0 \% & 4.7 d & \cellcolor[rgb]{ .886,  .937,  .855}79.2 kWh &    &    & 66.6 \% & 5.3 d & \cellcolor[rgb]{ .886,  .937,  .855}113.4 kWh \\
    CIFAR-100 & Random fc &    & 65.4 \% & 6.4 d & \cellcolor[rgb]{ .886,  .937,  .855}89.8 kWh &    &    & 65.7 \% & 6.5 d & \cellcolor[rgb]{ .886,  .937,  .855}97.8 kWh \\
    DenseNet-121 & Oort &    & 65.9 \% & 4.7 d & \cellcolor[rgb]{ .886,  .937,  .855}96.1 kWh &    &    & 65.9 \% & 6.5 d & \cellcolor[rgb]{ .886,  .937,  .855}130.9 kWh \\
       & Oort 1.3n &    & 66.4 \% & 4.5 d & \cellcolor[rgb]{ .886,  .937,  .855}103.8 kWh &    &    & 66.4 \% & 5.4 d & \cellcolor[rgb]{ .886,  .937,  .855}138.7 kWh \\
       & Oort fc &    & 65.8 \% & 5.3 d & \cellcolor[rgb]{ .886,  .937,  .855}102.4 kWh &    &    & 66.1 \% & 6.4 d & \cellcolor[rgb]{ .886,  .937,  .855}126.7 kWh \\
       & \textbf{FedZero} &    & \textbf{66.8 \%} & \textbf{3.6 d} & \cellcolor[rgb]{ .886,  .937,  .855}\textbf{70.4 kWh} &    &    & \textbf{66.5 \%} & \textbf{4.5 d} & \cellcolor[rgb]{ .886,  .937,  .855}\textbf{96.4 kWh} \\
    \midrule
       & Upper bound & \multirow{8}[2]{*}{62.4 \%} & 64.1 \% & 1.4 d & \cellcolor[rgb]{ .851,  .851,  .851}81.3 kWh &    & \multirow{8}[2]{*}{62.8 \%} & 64.1 \% & 1.8 d & \cellcolor[rgb]{ .851,  .851,  .851}105.7 kWh \\
       & Random &    & 62.4 \% & 6.7 d & \cellcolor[rgb]{ .886,  .937,  .855}99.8 kWh &    &    & 62.8 \% & 5.7 d & \cellcolor[rgb]{ .886,  .937,  .855}92.0 kWh \\
       & Random 1.3n &    & 63.1 \% & 5.6 d & \cellcolor[rgb]{ .886,  .937,  .855}109.6 kWh &    &    & 63.3 \% & 3.7 d & \cellcolor[rgb]{ .886,  .937,  .855}86.0 kWh \\
    Tiny ImageNet & Random fc &    & 63.0 \% & 5.6 d & \cellcolor[rgb]{ .886,  .937,  .855}96.4 kWh &    &    & 63.1 \% & 6.5 d & \cellcolor[rgb]{ .886,  .937,  .855}102.1 kWh \\
    EfficientNet-B1 & Oort &    & 63.3 \% & 3.8 d & \cellcolor[rgb]{ .886,  .937,  .855}88.1 kWh &    &    & 62.7 \% & -  & \cellcolor[rgb]{ .886,  .937,  .855}- \\
       & Oort 1.3n &    & 63.2 \% & 3.3 d & \cellcolor[rgb]{ .886,  .937,  .855}90.2 kWh &    &    & 63.5 \% & 3.7 d & \cellcolor[rgb]{ .886,  .937,  .855}112.4 kWh \\
       & Oort fc &    & 63.1 \% & 3.9 d & \cellcolor[rgb]{ .886,  .937,  .855}89.0 kWh &    &    & 59.7 \% & -  & \cellcolor[rgb]{ .886,  .937,  .855}- \\
       & \textbf{FedZero} &    & \textbf{63.7 \%} & \textbf{2.8 d} & \cellcolor[rgb]{ .886,  .937,  .855}\textbf{64.8 kWh} &    &    & \textbf{63.8 \%} & \textbf{3.4 d} & \cellcolor[rgb]{ .886,  .937,  .855}\textbf{76.6 kWh} \\
    \midrule
       & Upper bound & \multirow{8}[2]{*}{50.4 \%} & 53.3 \% & 1.4 d & \cellcolor[rgb]{ .851,  .851,  .851}82.5 kWh &    & \multirow{8}[2]{*}{50.9 \%} & 53.3 \% & 1.8 d & \cellcolor[rgb]{ .851,  .851,  .851}104.2 kWh \\
       & Random &    & 50.4 \% & 6.7 d & \cellcolor[rgb]{ .886,  .937,  .855}131.5 kWh &    &    & 50.9 \% & 5.7 d & \cellcolor[rgb]{ .886,  .937,  .855}93.2 kWh \\
       & Random 1.3n &    & 50.7 \% & 4.6 d & \cellcolor[rgb]{ .886,  .937,  .855}97.9 kWh &    &    & 51.5 \% & 4.5 d & \cellcolor[rgb]{ .886,  .937,  .855}90.0 kWh \\
    Shakespeare & Random fc &    & 52.0 \% & 2.8 d & \cellcolor[rgb]{ .886,  .937,  .855}57.3 kWh &    &    & 52.1 \% & 3.7 d & \cellcolor[rgb]{ .886,  .937,  .855}59.2 kWh \\
    LSTM & Oort &    & 50.2 \% & -  & \cellcolor[rgb]{ .886,  .937,  .855}- &    &    & 50.7 \% & -  & \cellcolor[rgb]{ .886,  .937,  .855}- \\
       & Oort 1.3n &    & 50.4 \% & -  & \cellcolor[rgb]{ .886,  .937,  .855}- &    &    & 51.7 \% & 4.5 d & \cellcolor[rgb]{ .886,  .937,  .855}95.4 kWh \\
       & Oort fc &    & 50.5 \% & 6.7 d & \cellcolor[rgb]{ .886,  .937,  .855}157.4 kWh &    &    & 50.6 \% & -  & \cellcolor[rgb]{ .886,  .937,  .855}- \\
       & \textbf{FedZero} &    & \textbf{53.1 \%} & \textbf{1.8 d} & \cellcolor[rgb]{ .886,  .937,  .855}\textbf{40.0 kWh} &    &    & \textbf{53.1 \%} & \textbf{2.3 d} & \cellcolor[rgb]{ .886,  .937,  .855}\textbf{42.8 kWh} \\
    \midrule
       & Upper bound & \multirow{8}[2]{*}{83.6 \%} & 87.9 \% & 2.7 d & \cellcolor[rgb]{ .851,  .851,  .851}105.6 kWh &    & \multirow{8}[2]{*}{82.8 \%} & 87.9 \% & 2.3 d & \cellcolor[rgb]{ .851,  .851,  .851}91.0 kWh \\
       & Random &    & 83.6 \% & 7.0 d & \cellcolor[rgb]{ .886,  .937,  .855}102.4 kWh &    &    & 82.8 \% & 6.7 d & \cellcolor[rgb]{ .886,  .937,  .855}84.2 kWh \\
       & Random 1.3n &    & 85.2 \% & 5.5 d & \cellcolor[rgb]{ .886,  .937,  .855}103.5 kWh &    &    & 85.1 \% & 4.3 d & \cellcolor[rgb]{ .886,  .937,  .855}80.8 kWh \\
    Google Speech & Random fc &    & 85.0 \% & 5.7 d & \cellcolor[rgb]{ .886,  .937,  .855}95.6 kWh &    &    & 83.0 \% & 6.7 d & \cellcolor[rgb]{ .886,  .937,  .855}83.8 kWh \\
    KWT-1 & Oort &    & 86.1 \% & 4.5 d & \cellcolor[rgb]{ .886,  .937,  .855}99.0 kWh &    &    & 86.2 \% & 3.7 d & \cellcolor[rgb]{ .886,  .937,  .855}78.2 kWh \\
       & Oort 1.3n &    & 86.8 \% & 3.9 d & \cellcolor[rgb]{ .886,  .937,  .855}103.5 kWh &    &    & 86.4 \% & 3.4 d & \cellcolor[rgb]{ .886,  .937,  .855}85.0 kWh \\
       & Oort fc &    & 87.0 \% & 3.8 d & \cellcolor[rgb]{ .886,  .937,  .855}86.2 kWh &    &    & 85.6 \% & 3.7 d & \cellcolor[rgb]{ .886,  .937,  .855}78.1 kWh \\
       & \textbf{FedZero} &    & \textbf{87.3 \%} & \textbf{3.6 d} & \cellcolor[rgb]{ .886,  .937,  .855}\textbf{77.4 kWh} &    &    & \textbf{87.5 \%} & \textbf{2.6 d} & \cellcolor[rgb]{ .886,  .937,  .855}\textbf{65.6 kWh} \\
    \bottomrule
    \end{tabular}%

\end{document}